\documentclass[journal]{IEEEtran}
\ifCLASSINFOpdf
\else
\fi
\usepackage{graphicx}
\usepackage{url}
\usepackage{amsmath,amssymb} 
\usepackage[ruled,vlined]{algorithm2e}
\usepackage{xcolor}

\begin{document}
%
\title{Knowledge Guided Disambiguation for Large-Scale Scene Classification with Multi-Resolution CNNs}
%
%
%

\author{Limin~Wang,
        Sheng~Guo,
        Weilin~Huang, ~\IEEEmembership{Member,~IEEE,}
        Yuanjun~Xiong,
        and~Yu~Qiao,~\IEEEmembership{Senior~Member,~IEEE}

\thanks{This work was supported in part by National Key Research and Development Program of China (2016YFC1400704), National Natural Science Foundation of China (U1613211, 61633021, 61503367), External Cooperation Program of BIC, Chinese Academy of Sciences (172644KYSB20160033), and Science and Technology Planning Project of Guangdong Province (2015A030310289).}
\thanks{L. Wang is with the Computer Vision Laboratory, ETH Zurich, Zurich, Switzerland (07wanglimin@gmail.com).} 
\thanks{S. Guo is with the Shenzhen Institutes of Advanced Technology, Chinese Academy of Sciences, Shenzhen, China. He is also with the Shenzhen College of Advanced Technology, University of Chinese Academy of Sciences, Beijing, China (e-mail: guosheng1001@gmail.com).}
\thanks{W. Huang is with the Shenzhen Institutes of Advanced Technology, Chinese Academy of Sciences, Shenzhen, China. He is also with the Visual Geometry Group, University of Oxford, UK (e-mail: wl.huang@siat.ac.cn).}
\thanks{Y. Xiong is with the Department of Information Engineering, The Chinese University of Hong Kong, Hong Kong (e-mail: bitxiong@gmail.com).}
\thanks{Y. Qiao is with the Guangdong Key Laboratory of Computer Vision and Virtual Reality, the Shenzhen Institutes of Advanced Technology, Chinese Academy of Sciences, Shenzhen, China. He is also with The Chinese University of Hong Kong, Hong Kong (yu.qiao@siat.ac.cn).}
}

\maketitle

\begin{abstract}
Convolutional Neural Networks (CNNs) have made remarkable progress on scene recognition, partially due to these recent large-scale scene datasets, such as the Places and Places2. Scene categories are often defined by multi-level information, including local objects, global layout, and background environment, thus leading to large intra-class variations. In addition, with the increasing number of scene categories, label ambiguity has become another crucial issue in large-scale classification. 
This paper focuses on large-scale scene recognition and makes two major contributions to tackle these issues. First, we propose a multi-resolution CNN architecture that captures visual content and structure at multiple levels. The multi-resolution CNNs are composed of coarse resolution CNNs and fine resolution CNNs, which are complementary to each other. Second, we design two knowledge guided disambiguation techniques to deal with the problem of label ambiguity. (i) We exploit the knowledge from the confusion matrix computed on validation data to merge ambiguous classes into a super category. (ii) We utilize the knowledge of extra networks to produce a soft label for each image. Then the super categories or soft labels are employed to guide CNN training on the Places2. We conduct extensive experiments on three large-scale image datasets (ImageNet, Places, and Places2), demonstrating the effectiveness of our approach. Furthermore, our method takes part in two major scene recognition challenges, and achieves the second place at the Places2 challenge in ILSVRC 2015, and the first place at the LSUN challenge in CVPR 2016. Finally, we directly test the learned representations on other scene benchmarks, and obtain the new state-of-the-art results on the MIT Indoor67 (86.7\%) and SUN397 (72.0\%). We release the code and models at~\url{https://github.com/wanglimin/MRCNN-Scene-Recognition}.
\end{abstract}

\begin{IEEEkeywords}
Scene recognition, large-scale recognition, multi-resolutions, disambiguation, Convolutional Neural Network.
\end{IEEEkeywords}

%
\IEEEpeerreviewmaketitle

\section{Introduction}

\begin{figure}[t]
\includegraphics[width=\linewidth]{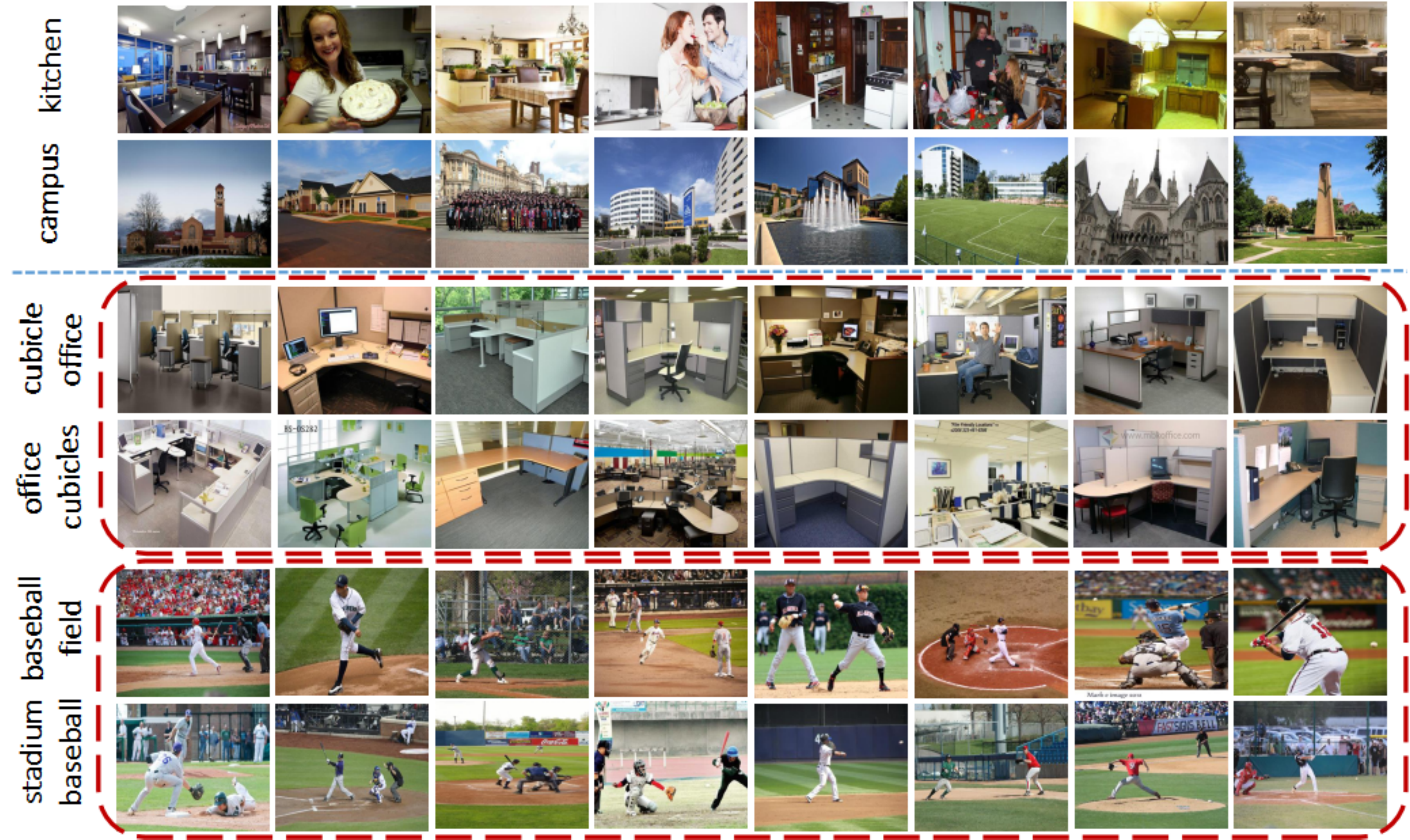}
\caption{Image examples from the Places2 dataset. \textbf{Top Row}:  we show images from two separate scene classes (i.e., \textit{Kitchen} and \textit{Campus}). We notice that large intra-class variations are contained in these images.  \textbf{Bottom Row}: we give two pairs of scene categories (i.e., \textit{Cubicle office} and \textit{Office cubicles},  \textit{Baseball field} and \textit{Stadium baseball})). As can be found, images from these pairs of ambiguous categories are highly confused.}
\label{fig:ex}
\end{figure}

\IEEEPARstart{S}{cene} recognition is a fundamental problem in computer vision, and has received increasing attention in the past few years~\cite{LiPT05,WuR11,ZhangZS14,ZuoWSZYJ14,GongWGL14,XieWGZT14,YangR15,ShenLH15,ZuoSWLWWC16,guo2016locally,WangWWZ016}. Scene recognition not only provides rich semantic information of global structure~\cite{OlivaT01}, but also yields meaningful context that facilitates other related vision tasks, such as object detection~\cite{Torralba03,FelzenszwalbGMR10,WangWLC11}, event recognition~\cite{wang2015object,XiongZLT15,WangWQG16}, and action recognition~\cite{Wang0T16,ZhangWCDL16,WangQT14}. In general, it is assumed that scene is composed of specific objects arranged in a certain layout, so that scene categories are often defined by multi-level information, including local objects, global layout, background environments, and possible interactions between them. Compared with object categories, the concept of scene is more subjective and complicated, so that there may not exist consensus on how to define an environment category. These pose main challenges for developing an effective and robust algorithm that is able to compute all these multi-level information from images.

Recently, large-scale scene datasets (e.g., the Places~\cite{ZhouLXTO14} and Places2~\cite{Places2}) have been introduced to advance the research on scene understanding,  allowing to train powerful convolutional neural networks (CNNs)~\cite{lecun-98} for scene classification. These large-scale datasets consist of a rich scene taxonomy, which includes rich categories to cover the diverse visual environments of our daily experience. With these scene category information, scene keywords could be sent to image search engines (e.g., Google Images, Bing Images or Flicker), where millions of images can be downloaded, and then be further sent to Amazon Mechanical Turk for manual annotation. However, as the number of classes grows rapidly, these visual categories start to overlap with each other. Thus there may exist label ambiguity among these scene classes. As shown in Figure~\ref{fig:ex}, images in \emph{cubicle office} and  \emph{office cubicles}  categories are easily confused with each other, similar ambiguities happen in \emph{baseball field} and \emph{stadium baseball}. Partially due to this reason, even the human top1 error rate is still relatively high on the SUN397 dataset (around 30\%)~\cite{XiaoHEOT10}.

Due to inherent uncertainty of scene concepts and increasing overlap among different categories, it is challenging to conduct scene recognition on large-scale datasets containing hundreds of classes and millions of images. Specifically, current large-scale datasets pose two major challenges for scene classification, namely \emph{visual inconsistence} and \emph{label ambiguity}.
\begin{itemize}
 \item For {\bf visual inconsistence}, we refer to the fact that there exist large variations among  images from the same scene category. Since it is difficult to define scene categories objectively, natural images are annotated according to annotators' subjective experiences when a dataset is created. This naturally leads to strong diversity on large-scale scene datasets. For instance, images in \emph{kitchen} category contain significantly diverse context and appearance, ranging from a whole room with many cooking wares to a single people with food, as shown in Figure~\ref{fig:ex}.
 \item For {\bf label ambiguity}, we argue that some scene categories may share similar visual appearance, and could be easily confused with others. As the number of scene classes increases, the inter-category overlaps can become non-negligible. For example, as shown in Figure \ref{fig:ex}, the {\em baseball field} category is very similar to the {\em stadium baseball}, and they both contain identical representative objects, such as track and people.
\end{itemize}

These challenges motivate us to develop an effective multi-resolution disambiguation model for large-scale scene classification, by making two major contributions: (1) {\em we propose a multi-resolution convolutional architecture to capture multi-level visual cues of different scales}; (2) {\em We introduce knowledge guided strategies to effectively disambiguate similar scene categories}. {\bf First}, to deal with the problem of visual inconsistence (i.e., large intra-class variations), we come up with a multi-resolution CNN framework, where CNNs at coarse resolution are able to capture global structure or large-scale objects, while CNNs at fine resolution are capable of describing local detailed information of fine-scale objects. Intuitively, multi-resolution CNNs combine complementary visual cues of multi-level concepts, allowing them to tackle the issue of large intra-class variations efficiently. {\bf Second}, for the challenge of label ambiguity (i.e., small inter-class variations), we propose to reorganize the semantic scene space to release the difficulty of training CNNs, by exploiting extra knowledge. In particular, we design two methods with the assistance from confusion matrix computed on validation dataset and publicly available CNN models, respectively. In the first method, we investigate the correlation of different classes and progressively merge similar categories into a super category. In the second one, we use the outputs of extra CNN models to generate new labels. These two methods essentially utilize extra knowledge to produce new labels for training images. These new labels are able to guide the CNN to a better optimization and reduce the effect of over-fitting.

To verify the effectiveness of our method, we choose the successful BN-Inception architecture~\cite{IoffeS15} as our basic network structure, and demonstrate the advantages of multi-resolution CNNs and knowledge guided disambiguation strategies on a number of benchmarks. More specifically, we first conduct experiments on three large-scale image recognition datasets, including the ImageNet~\cite{DengDSLL009}, Places~\cite{ZhouLXTO14}, and Places2~\cite{Places2}, where our method obtains highly competitive performance. Then, we further apply the proposed framework on two high-impact scene recognition challenges, namely the Places2 challenge (held in ImangeNet large scale visual recognition challenge~\cite{ILSVRC15}) and the large-scale scene understanding (LSUN) challenge in CVPR 2016. Our team secures the second place at the Places2 challenge 2015 and the first place at the LSUN challenge 2016. Furthermore, we evaluate the generalization ability of our learned models by testing them directly on the MIT Indoor67~\cite{QuattoniT09} and SUN397~\cite{XiaoHEOT10} benchmarks, with new state-of-the-art performance achieved. Finally, we present several failure cases by our models to highlight existing challenges for scene recognition, and discuss possible research directions in the future.

The rest of the paper is organized as follows. In Section \ref{sec:rw}, we review related works from aspects of scene recognition, deep networks for image recognition, and knowledge transferring. Section \ref{sec:mr_cnn} introduces the architecture of multi-resolution convolutional neural networks. In Section \ref{sec:kd}, we develop two types of knowledge guided disambiguation strategies to improve the performance of scene recognition. We report experimental results and analyze different aspects of our method in Section \ref{sec:con}. Finally, we conclude the paper in Section \ref{sec:con}.

\section{Related Work}
\label{sec:rw}

In this section, we briefly review previous works related to our method, and clarify the difference between them. Specifically, we present related studies from three aspects: (1) scene recognition, (2) deep networks for image recognition, (3) multi-scale representation, and (4) knowledge transfer.

\textbf{Scene recognition.} The problem of scene recognition has been extensively studied in previous works. For example, Lazebnik \emph{et al.}~\cite{LazebnikSP06} proposed spatial pyramid matching (SPM) to incorporate spatial layout into bag-of-word (BoW) representation for scene recognition. Partizi \emph{et al.}~\cite{PariziOF12} designed a reconfigurable version of SPM, which associated different BoW representations with various image regions. The standard deformable part model (DPM) \cite{FelzenszwalbGMR10} was extended to scene recognition by Pandey \emph{et al.} \cite{PandeyL11}. Quattoni \emph{et al.} \cite{QuattoniT09} studied the problem of indoor scene recognition by modeling the spatial layout of scene components. Mid-level discriminative patches or parts were discovered and identified  for scene recognition in \cite{SinghGE12,JunejaVJZ13}. Recently, deep convolutional networks have been exploited for scene classification by Zhou \emph{et al.}~\cite{ZhouLXTO14}, where they introduced a large-scale Places dataset and advanced the state of the art of scene recognition by a large margin. After this, they introduced a more challenging dataset~\cite{Places2} with more categories and images, coined as Places2. 

Our paper differs from these previous works from two aspects: (1) We tackle scene recognition problem with a much larger scale database, where new problems, such as large visual inconsistent and  significant category ambiguity, are raised. These make our problem more challenging than all previous ones on this task. (2) We design a multi-resolution architecture and propose a knowledge guided disambiguation strategy that effectively handle these new problems. Large-scale problem is the fundamental challenge in computer vision, we provide a high-performance model on such a challenging dataset, setting our work apart from all previous ones on scene recognition.

\textbf{Deep networks for image recognition.} Since the remarkable progress made by AlexNet \cite{KrizhevskySH12} on ILSVRC 2012, great efforts have been devoted to the problem of image recognition with various deep learning techniques \cite{ZeilerF14,HeZR014,SimonyanZ14a,SzegedyLJSRAEVR15,IoffeS15,HeZRS15a,ShenLH15,Szegedy_2016_CVPR,HeZRS15}. A majority of these works focused on designing deeper network architectures, such as VGGNet \cite{SimonyanZ14a}, Inception Network \cite{SzegedyLJSRAEVR15,Szegedy_2016_CVPR}, and ResNet \cite{HeZRS15} which finally contains hundreds of layers. Meanwhile, several regularization techniques and data augmentations have been designed to reduce the over-fitting effect of the network, such as dropout \cite{KrizhevskySH12}, smaller convolutional kernel size \cite{ZeilerF14,SimonyanZ14a}, and multi-scale cropping \cite{SimonyanZ14a}. In addition, several optimization techniques have been also proposed to reduce the difficulty of training networks, so as to improve recognition performance, such as Batch Normalization (BN) \cite{IoffeS15} and Relay Back Propagation \cite{ShenLH15}. 

These works focused on general aspect of applying deep networks for image classification, in particular for object recognition, without considering the specifics of scene recognition problem. As discussed, sense categories more complicated than object classes. They are defined by multi-level visual concepts, ranging from local objects, global arrangements, to dynamic interactions between them.  Complementary to previous works on object classification, we conduct a dedicated study on the difficulties of scene recognition, and accordingly come up with two new solutions that address the crucial issues existed in large-scale scene recognition. We propose a multi-resolution architecture to capture visual information from multi-visual concepts, and wish to remedy the visual inconsistence problem. In addition, we design a knowledge guided disambiguation mechanism that effectively handles the issue of label ambiguity, which is an another major challenge for this task.

\textbf{Multi-scale representation.} The idea of multi-scale or multi-resolution representations has been widely studied in the computer vision research. First, the multi-scale cropping was first adopted for network training by the VGGNet~\cite{SimonyanZ14a} and then commonly used by the following deep networks, such as ResNet~\cite{HeZRS15} and Inception V3~\cite{Szegedy_2016_CVPR}. The multi-scale representation have been also exploited in variety of tasks, such as fine-grained recognition~\cite{ZhangWWCLND16}, scene recognition~\cite{WangWWZ016}, and so on. Zhang {\em et al.}~\cite{ZhangWWCLND16} generated multi-scale part proposals and these proposals yielded the multi-scale image representation for fine-grained categorization. Wang {\em et al.}~\cite{WangWWZ016} extracted multi-scale patches for PatchNet modeling and VSAD representations.

Different from these multi-scale cropping and multi-scale representation, our multi-resolution architecture captures multi-level information from different resolutions with distinct input image sizes and network architectures, while those previous works all rely on a single input size and network architecture. Meanwhile, the multi-scale cropping (scale jittering) is complementary to our multi-resolution architecture and we also exploit this data augmentation in our training method.

\textbf{Knowledge transfer.} Knowledge distillation or knowledge transferring between different CNN models is becoming an active topic recently \cite{HintonVD15,RomeroBKCGB14,GuptaHM15,TzengHDS15,ZhangWWQW16}. The basic idea of using the outputs of one network as an associated supervision signal to train a different model was invented by Bucila \emph{et al.} \cite{BucilaCN06}. Recently, Hinton \emph{et al.} \cite{HintonVD15} adopted this technique to compress model ensembles into a smaller one for fast deployment. Similarly, Romero \emph{et al.} \cite{RomeroBKCGB14} utilized this approach to help train a deeper network in multiple stages. Tzeng \emph{et al.} \cite{TzengHDS15} explored this method to the problem of domain adaption for object recognition. Gupta \emph{et al.} \cite{GuptaHM15} proposed to distill knowledge across different modalities, and used RGB CNN models to guide the training of CNNs for depth maps or optical flow field. Zhang \emph{et al.} \cite{ZhangWWQW16} developed a knowledge transfer technique to exploits soft codes of flow CNNs to assist the training of motion vector CNNs, with a goal of real-time action recognition from videos.

Our utilization of soft codes as an extra supervision signal differs from these methods mainly from two points: (1) we conduct knowledge transfer crossing different visual tasks (e.g., object recognition vs. scene recognition), while previous methods mostly focused on the same task; (2) we exploit these soft codes to circumvent label ambiguity problem existed in large-scale scene recognition.

\begin{figure*}[ht]
\includegraphics[width=\linewidth]{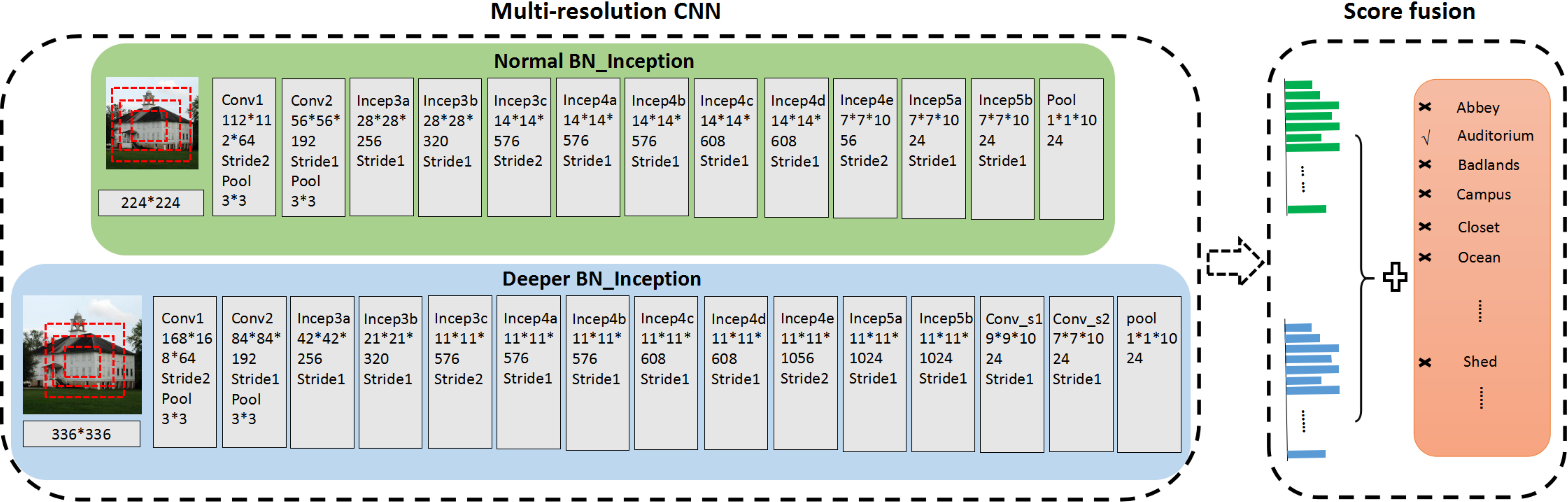}
\caption{\textbf{Multi-resolution CNN.} We propose a multi-resolution architecture, which is composed of coarse resolution CNNs (normal bn-inception) and fine resolution CNNs (deeper bn-inception). The coarse resolution CNNs capture visual structure at a large scale, while fine resolution CNNs describe visual pattern at a relatively smaller scale. The receptive fields (red boxes) of two CNNs correspond to the regions of different scales, allowing their prediction scores to be complementary.}
\label{fig:mrcnn}
\end{figure*}

\section{Multi-Resolution Convolutional Neural Networks}
\label{sec:mr_cnn}

Generally, a visual scene can be defined as a view that objects and other semantic surfaces are arranged in a meaningful way \cite{Oliva09}. Scenes contain semantic components arranged in a spatial layout which can be observed at a variety of spatial scales, e.g., the up-close view of an office desk or the view of the entire office. Therefore, when building computational models to perform scene recognition, we need to consider this multi-scale property of scene images. Specifically, in this section, we first describe the basic network structure used in our exploration, and then present the framework of multi-resolution CNN.

\subsection{Basic network structures}
Deep convolutional networks have witnessed great successes in image classification and many powerful network architectures have been developed, such as AlexNet \cite{KrizhevskySH12}, GoogLeNet \cite{SzegedyLJSRAEVR15}, VGGNet \cite{SimonyanZ14a}, and ResNet \cite{HeZRS15}. As the dataset size of Places2 is much larger than that of ImageNet, we need to trade off between recognition performance and computational cost when building our network structure. In our experiments, we employ the inception architecture with batch normalization \cite{IoffeS15} (bn-inception) as our network structure. In addition to its efficiency, the inception architecture leverages the idea of multi-scale processing in its inception modules, making it naturally suitable for building scene recognition networks.

As shown in Figure 2, the original bn-inception architecture starts with two convolutional layers and max pooling layers which transform a $224 \times 224$ input image into $28 \times 28$ feature maps. The small size of feature maps allows for fast processing in the subsequent ten inception layers,  two of which have stride of 2 and the rest have stride of 1. This results in $7 \times 7$ feature maps, and a global average pooling is used to aggregate these activations across spatial dimensions. Batch Normalization (BN) is applied to the activations of convolutional layers, following by the Rectified Linear Unit (ReLU) for non-linearity.

\begin{figure*}[t]
\includegraphics[width=\linewidth]{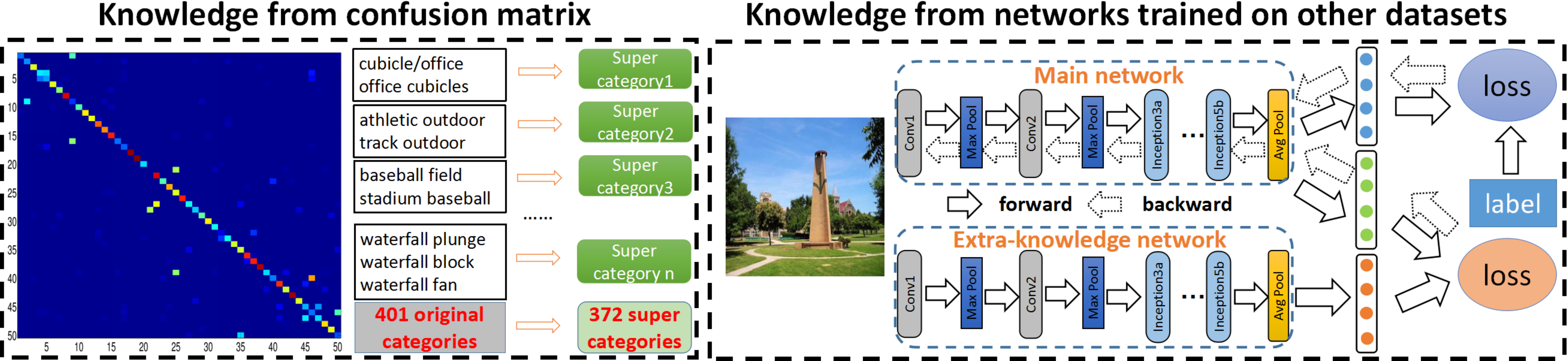}
\caption{\textbf{Knowledge guided disambiguation.} We propose two knowledge guided disambiguation methods to deal with the problem of label ambiguity. First (in left), we utilize the knowledge of confusion matrix to merge ambiguous scene classes into a super category, and re-train our CNNs on these re-labeled data. Second (in right), we exploit the knowledge of extra networks (trained on a different dataset) to provide additional supervised information for each image (soft label), which are used to guide the CNN training in multi-task learning framework.}
\label{fig:kd}
\end{figure*}

\subsection{Two-resolution architectures}
The proposed Multi-Resolution CNNs are decomposed into fine resolution and coarse resolution components. The coarse resolution CNNs are the same with the normal bn-inception described in previous subsection, while the fine resolution CNNs share a similar but deeper architecture.

\textbf{Coarse resolution CNNs} operate on image regions of size $224 \times 224$, and contain totally $13$ layers with weights. The network structure of coarse resolution CNNs is referred as \emph{normal bn-inception}, since it has the same structure as the original one in \cite{IoffeS15}. It captures visual appearance and structure at a relatively coarse resolution, focusing on describing global arrangements or objects at larger scale. 
However, the coarse resolution CNNs may discard local details, such as those fine-scale objects, which are important cues to discriminate sense categories. A powerful scene network should be able to describe multi-level visual concepts, so that it requires to capture visual content in a finer resolution where local detail information is enhanced.

\textbf{Fine resolution CNNs} are developed for high-resolution images of $384 \times 384$, and process on image regions of $336 \times 336$. By taking a larger image as input, the depth of the network can be increased, allowing us to design a new model with increasing capability. By trading off model speed and network capacity, we add three extra convolutional layers on top of the inception layers, as illustrated in Figure \ref{fig:mrcnn}. For these newly-added convolutional layers, the pad sizes are set as zeros, so as to keep the resulting feature map as the same size of $7 \times 7$ before the global average pooling. We refer this network structure of fine resolution CNN as \emph{deeper bn-inception}, which aims to describe image information and structure at finer scale, allowing it to capture meaningful  local details.

Our two-resolution CNNs take different resolution images as input, so that their receptive fields of the corresponding layers describe different-size regions in the original image, as illustrated in Figure \ref{fig:mrcnn}. They are designed to describe objects at different scales for scene understanding. Therefore, the prediction scores of our two-resolution models are complementary to each other, by computing an arithmetic average of them. 

\textbf{Extension to multi-resolution CNNs.} The above the description is about two-resolution CNNs (i.e., learning CNNs from two resolutions: $256 \times 256$ and $384 \times 384$), and the idea could be easily extended to multi-resolution CNNs. In practice, we can train CNNs from multiple resolutions, and the CNNs learned from finer resolution are expected to be equipped richer capacity of modeling visual information and structure. Meanwhile, CNNs trained from more resolution complement each other more effectively and is hoped to improve the final recognition performance greatly. In experiment, we conduct experiments with four resolutions (128, 256, 384, 512) to extensively study the influence of image resolution on the recognition performance and fully reveal the modeling capacity of our multi-resolution CNN framework.

\textbf{Discussion.} Although sharing similar ideas with common multi-scale training strategy \cite{SimonyanZ14a}, the proposed multi-resolution CNNs differ from it distinctly. The network input image sizes are different in our two-resolution architectures ($224 \times 224$ and $336 \times 336$), while multi-scale training in \cite{SimonyanZ14a} only uses a single image scale,  $224 \times 224$. This allow us to design two distinct network structures (the \textit{bn-inception} and \textit{deeper bn-inception}) with enhanced model capability, which are capable of handling different image scales. The conventional multi-scale training simply uses a single network structure. Thanks to these differences, the proposed multi-resolution architecture is more suitable to capture different level visual information for scene understanding. Moreover, the multi-resolution architecture is complementary to multi-scale training, and can be easily combined with it as stated in next paragraph.

\textbf{Training.} The training of multi-resolution CNNs are performed for each resolution independently. We train each CNNs according to common setup of  \cite{KrizhevskySH12,SimonyanZ14a}. We use the mini-batch stochastic gradient descent algorithm to learn the network weights, where the batch size is set as 256 and momentum set to 0.9. The learning rate is initialized as $0.1$ and decreases according to a fixed schedule determined by the dataset size and specified in Section \ref{sec:exp}. Concerning data augmentation, the training images are resized as $N \times N$, where $N$ is set as $256$ for the bn-inception, and $384$ for the deeper bn-inception. Then, we randomly crop a $w \times h$ region at a set of fixed positions, where the cropped width $w$ and height $h$ are picked from $\{N, 0.875N, 0.75N, 0.625N, 0.5N\}$. Then these cropped regions are resized as $M \times M$ for network training, where $M$ depends on the image resolution $N$ and is set as $0.875N$. Meanwhile, these crops undergo a horizontal flipping randomly. Our proposed cropping strategy is an efficient way to implement the scale jittering~\cite{SimonyanZ14a}.

\section{Knowledge Guided Disambiguation}
\label{sec:kd}

\begin{figure*}
\includegraphics[width=\textwidth]{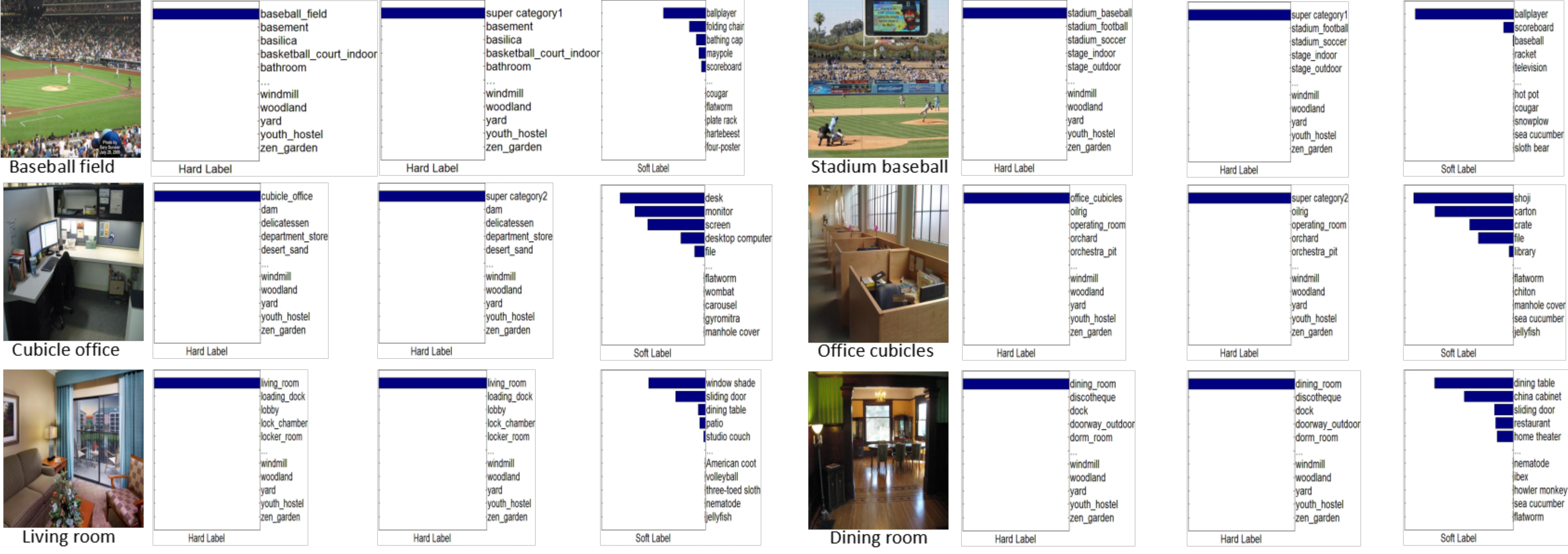}
\caption{\textbf{Hard and soft labels.} Several image examples with ground truth from the Places2 dataset. First (left column), we show the original hard labels provided by the dataset. Second (middle column), the hard labels are shown after merging visually ambiguous classes (by our first disambiguation approach). As can be found, the classes of \textit{baseball field} and \textit{stadium baseball} are merged into super category 1, while the classes of \textit{cubicle office} and \textit{office cubicles} are merged into super category 2. Finally (right column),  we provide the soft labels produced by an extra network (by our second disambiguation approach), where scene content is described by the distribution over common objects from existing ImangeNet CNNs (knowledge models).}
\label{fig:label}
\end{figure*}

As analyzed above, many scene categories may overlap with others in large-scale datasets, such as Places2~\cite{Places2}. The increasing number of scene categories causes the problem of label ambiguity, which makes the training of multi-resolution CNNs more challenging. In this section, we propose two simple yet effective methods to handle the issue of label ambiguity by exploiting extra knowledge. Specifically, we first introduce the method of utilizing knowledge from confusion matrix.  Then we propose the second one which resorts to knowledge from extra networks.

\subsection{Knowledge from confusion matrix}

\begin{algorithm}[t]
  \SetAlgoLined
  \KwData{Similarity matrix $\mathbf{S}$, threshold: $\tau$.}
  \KwResult{Merged classes: $\mathcal{S}$.}
  - Initialization: $\mathcal{S} =  \leftarrow \{S_1,S_2,\cdots, S_N\}$. \\
  \While{$max(\mathbf{S}) < \tau$}{
    1. Pick the  maximum of similarity: $(i,j)^* \leftarrow \mathrm{arg}\max_{i,j} \mathbf{S}_{ij}$ \\
    2. Merge the $i^{*th}$ and $j^{*th}$ classes into a single class : $\mathcal{S} = \mathcal{S} - \{S_{i^*}\} - \{S_{j^*}\} + \{(S_{i^*}, S_{j^*})\}$ \\
    3. Update the similarity matrix by deleting $i^{*th}$ and $j^{*th}$ rows and columns and adding a new row and column defined as $\frac{1}{2}(\mathbf{S}_i + \mathbf{S}_j)$
    }
  - Return merged classes: $\mathcal{S}$.
  \caption{Merge similar classes into super category}
  \label{alg:merge}
\end{algorithm}

As the number of scene classes increases, the difference between scene categories becomes smaller, and some scene classes are easily confused with others from visual appearance. A natural way to relieve this problem is to re-organize scene class hierarchy, and merge those highly ambiguous ones into a super category. The key to merge them accurately is to define the similarity between categories. However, it is difficult to define this similarity (ambiguity) and merge them manually, which is highly subjective and extremely time-consuming on such a large-scale problem. Here we propose a simple yet effective approach that automatically merges visually ambiguous scene categories.

Specifically, we first train a deep model on the original training set of the Places2 which contains 401 classes. Then, we use the trained model to predict image categories on the validation set of the Places2. The confusion matrix is computed by using the predicted categories and ground-truth labels. This confusion matrix displays crossing errors between pairs of categories, which implicitly indicates the degree of ambiguity (similarity) between them. Hence, it is principle to employ this confusion matrix for calculating the pairwise similarities of scene classes. Formally, we define the similarity as follows:
\begin{equation}
\mathbf{S} = \frac{1}{2}(\mathbf{C} + \mathbf{C}^\top),
\end{equation}
where $\mathbf{C} \in \mathbb{R}^{N \times N}$ is the confusion matrix, $\mathbf{C}_{ij}$ represents the probability of classifying $i^{th}$ class as $j^{th}$ class,  the larger of this value indicates higher ambiguity between two categories. $N$ is the number of scene classes. The equation ensures the similarity measure is a symmetric metric.

This similarity measure computes underline relationships between categories, providing an important cue for constructing consistent super categories. To this end, we propose a bottom-up clustering algorithm that progressively merges ambiguous categories, as shown in Algorithm \ref{alg:merge}. At each iteration, we pick a pair of categories with the largest similarity, and merge them into a super category. Then we update the similarity matrix $\mathbf{S}$ accordingly, by deleting $i^{*th}$ and $j^{*th}$ rows and columns, and at the same time, adding a new row and column defined as $\frac{1}{2}(\mathbf{S}_{i^*} + \mathbf{S}_{j^*})$, where $\mathbf{S}_{i^*}$ denotes the $i^{*th}$ row vector of $\mathbf{S}$. This iteration repeats until there is no similarity value larger than $\tau$. 
At the end, all ambiguous classes are merged into a smaller number of categories, resulting in a more consistent category structure that greatly facilitates learning a better CNN. 

In current implementation, the original 401 scene classes in the Places401 are re-organized into 351, 372, and 386 super-categories by varying the threshold $\tau$ from 0.3, 0.5 to 0.7. In test phase, the re-trained model only predicts scores over super category labels. We transfer them to the original 401 categories, by equally dividing the probability of each super category into its sub categories. This simple strategy turns out to be effective in practice.

\subsection{Knowledge from extra networks}

Knowledge disambiguation by confusion matrix involves class-level re-organization, where we simply consider the similarity between whole classes, and merge them directly into a super category. However, this re-labeling (merging) strategy treats all images in a class equally, and ignores intra-class difference appeared in each single image. The confusion matrix defines category-level ambiguity (similarity) by computing the error rates with other classes, which means that only part of images from these visually ambiguous categories are classified incorrectly. It is more principle to involve image-level re-labeling strategy based on visual content of each image. Hence, in this subsection, we propose to exploit knowledge from extra networks to incorporate the visual information of each single image into this relabeling procedure.

However, it is prohibitively difficult to accomplish the work of image-level re-labeling manually in such a large-scale dataset. Furthermore, the category ambiguity may happen again, since it is challenging to define an objective re-labeling criteria. Fortunately, many CNN models trained on a relatively smaller and well-labeled dataset (e.g., the ImageNet~\cite{DengDSLL009} or Places~\cite{ZhouLXTO14}) are publicly available. These pre-trained models encode rich knowledge from different visual concepts, which is greatly helpful to guiding the image-level re-labeling procedure. They are powerful to extract high-level visual semantics from raw images. Therefore, we utilize these pre-trained models as knowledge networks to automatically assign soft labels to each image by directly using their outputs. 

Essentially, this soft label is a kind of distributed representation, which describes the scene content of each image with a distribution over the pre-trained class space. e.g.,  common object classes by using the ImageNet \cite{DengDSLL009}, or  a smaller subset of scene categories by using the Places \cite{ZhouLXTO14}. As shown in Figure \ref{fig:label}, for instance, the content of {\em dinning room} could be described by distribution of common objects, where objects such as {\em dinning table} and {\em door} may dominate this distribution. For another scene category such as {\em office}, the objects of {\em screen} and {\em desktop computer} may have high probability mass. Utilizing this soft label to represent image content exhibit two main advantages: (1) For visually ambiguous classes, they typically share similar visual elements such as objects and background. Hence they may have similar soft labels which encode the correlation of scene categories implicitly. (2) Compared with class-level re-labeling scheme, the soft label depends on single image content, so that it could be varied for different images in the same class. Normally, images from highly ambiguous classes may share similar but not identical soft labels. Hence, such soft labels are able to capture subtle difference between confused images, making them more informative and discriminative than hard labels.

In current implementation, we consider the complementarity between ground-truth hard labels and soft labels from knowledge networks, and design a multi-task learning framework that utilizes both labels to guide CNN training, as shown in Figure \ref{fig:kd}. Specifically, during the training procedure, our CNNs predict both the original hard labels and the soft labels simultaneously, by minimizing the following objective function:
\begin{equation}
\ell(D) = -(\sum_{\mathbf{I}_i \in D} \sum_{k=1}^{K_1} \mathbb{I}(y_i = k) \log p_{i,k} + \lambda \sum_{\mathbf{I}_i \in D} \sum_{k=1}^{K_2} q_{i,k} \log f_{i,k}),
\label{equ:mt}
\end{equation}
where $D$ denotes the training dataset. $\mathbf{I}_i$ is the $i^{th}$ image, $y_i$ is the ground-truth scene label (hard label), and $p_i$ is corresponding predicted scene label.
$f_i$ is its soft code (soft label) produced by extra knowledge network, and $q_i$ is the predicted soft code of image $\mathbf{I}_i$. $\lambda$ is a parameter balancing these two terms.  $K_1$ and $K_2$ are the dimensions of hard label and soft label, corresponding to the numbers of classes in main model and the knowledge model, respectively.

This multi-task objective function forces the training procedure to optimize the classification performance of original scene classification, and imitate the knowledge network at the same time. This multi-task learning framework is able to improve generalization ability by exploiting additional knowledge contained in extra networks as an inductive bias, and reduce the effect of over-fitting on the training dataset. For example, object concepts learned by the ImageNet pre-trained model provide important cues for distinguishing scene categories. As we shall see in Section \ref{sec:exp}, this framework further improves the performance of our proposed multi-resolution CNNs.

\section{Experiments}
\label{sec:exp}

In this section, we describe the experimental setting and report the performance of our proposed method on six scene benchmarks, including the ImageNet \cite{DengDSLL009}, Places \cite{ZhouLXTO14}, Places2 \cite{Places2}, LSUN \cite{YuZSSX15}, MIT Indoor67 \cite{QuattoniT09}, and SUN397 \cite{XiaoHEOT10} databases. We first describe these datasets and our implementation details. Then, we verify the effectiveness of multi-resolution CNNs by performing extensive experiments on three large-scale datasets. After this, we conduct experiments to explore the effect of knowledge guided disambiguation on the Places2. Furthermore, we report the performance of our method on two large-scale scene recognition challenges, namely the Places2 challenge in ILSVRC 2015, and the LSUN challenge in CVPR 2016. Meanwhile, we investigate generalization ability of our models, by directly testing the learned representations on the datasets of MIT Indoor67 \cite{QuattoniT09} and SUN397 \cite{XiaoHEOT10}. Finally, we present several failure examples by our methods, and discuss possible reasons.

\subsection{Large-scale datasets and implementation details}

\begin{table*}[t]
\begin{center}
\caption{Classification error of normal BN-Inception, deeper BN-Inception, and two-resolution CNN on the validation data of ImageNet-1k, Places205, Places401, and Places365.}
\label{tbl:mrcnn}
\begin{tabular}{|l|c|c|c|c|}
  \hline
  Method & ImageNet-1k (top1/top5)  & Places205 (top1/top5) & Places401 (top1/top5) & Places365 (top1/top5) \\
  \hline
  AlexNet~\cite{KrizhevskySH12} & 40.7\%/18.2\% & 50.0\%/- & 57.0\%/- & 46.8\%/17.1\% \\
  \hline
  VGGNet-16~\cite{SimonyanZ14a} & 27.0\%/8.8\% & 39.4\%/11.5\% & 52.4\%/- & 44.8\%/15.1\% \\
  \hline
  Normal BN-Inception &  24.7\%/7.2\% &  38.1\%/11.3\% & 48.8\%/17.4\% & 44.3\%/14.3\% \\
  \hline
  Deeper BN-Inception &  23.7\%/6.6\% &  37.8\%/10.7\% & 48.0\%/16.7\% & 44.0\%/14.0\% \\
  \hline
  Two-resolution CNN & 21.8\%/6.0\% & 36.4\%/10.4\% & 47.4\%/16.3\% & 42.8\%/13.2\% \\
  \hline
\end{tabular}
\end{center}
\end{table*}

\begin{table*}[t]
\begin{center}
\caption{Classification error of CNNs trained from different resolutions on the validation data of ImageNet-1k and Places365.}
\label{tbl:mrcnn1}
\begin{tabular}{|l|c|c|c|c|c|}
  \hline
  Resolution & $128 \times 128$ (top1/top5)  & $256 \times 256$ (top1/top5) & $384 \times 384$ (top1/top5) & $512 \times 512$ (top1/top5) & Fusion (top1/top5) \\
  \hline
  ImageNet-1k &  33.5\%/12.8\%& 24.7\%/7.2\% & 23.7\%/6.6\% &  23.5\%/6.6\% & 21.1\%/5.8\%\\
  \hline
  Places365 &  47.2\%/16.8\% & 44.3\%/14.3\% &  44.0\%/14.0\% & 43.5\%/13.8\% & 41.7\%/12.7\% \\
  \hline
\end{tabular}
\end{center}
\end{table*}

We first evaluate our method on three large-scale image classification datasets, namely ImageNet \cite{DengDSLL009}, Places \cite{ZhouLXTO14}, and Places2 \cite{Places2}. Results are reported and compared on their validation sets, since the ground-truth labels of their test sets are not available.

The ImageNet~\cite{DengDSLL009} is an object-centric dataset, and is the largest benchmark for object recognition and classification~\footnote{\url{http://image-net.org/}}. The dataset for ILSVRC 2012 contains 1,000 object categories (ImageNet-1k). The training data contains around 1,300,000 images from these object categories. There are 50,000 images for validation dataset and 100,000 images for testing. The evaluation measure is based on top5 error, where algorithms will produce a list of at most 5 object categories to match the ground truth.

The Places~\cite{ZhouLXTO14} is a large-scale scene-centric dataset~\footnote{\url{http://places.csail.mit.edu/}}, including 205 common scene categories (referred to as {\bf Places205}). The training dataset contains around 2,500,000 images from these categories.  In the training set, each scene category has the minimum 5,000 and maximum 15,000 images. The validation set contains 100 images per category (a total of 20,500 images), and the testing set includes 200 images per category (a total of 41,000 images). The evaluation criteria of the Places is also based on top5 error.

The Places2~\cite{Places2} is extended from the Places dataset, and probably the largest scene recognition dataset currently~\footnote{\url{http://places2.csail.mit.edu/}}. In total, the Places2 contains more than 10 million images comprising more than 400 unique scene categories. The dataset includes 5000 to 30,000 training images per class, which is consistent with real-world frequencies of occurrence. The dataset used in the Places2 challenge 2015 contains 401 scene categories ({\bf Places401}). The training dataset of the Places2 has around 8,100,000 images, while the validation set contains 50 images per category, and the testing set has 950 images per category. In consistent with our finding of label ambiguity, the latest version for the Places2 challenge 2016, has 365 scene categories, by merging similar scene categories into a single category ({\bf Places365}). The Places365 dataset has two training subsets: (1) {\em Places365-standard} has around 1.8 million training images and each category has around 5,000 images, and (2) {\em Places365-challenge} totally has around 8 million training images. We perform experiments and report results on the datasets of Places401 and Places365-standard. 

The training details of our proposed method on these three datasets are similar, as specified in Section \ref{sec:mr_cnn}.  The only difference is on the iteration numbers, due to the different sizes of training data on these datasets. Specifically, on the ImageNet and Places205 datasets, we decrease learning rate every 200,000 iterations and the whole training procedure stops at 750,000 iterations, while on the Places401 dataset, learning rate is decreased every 350,000 iterations and the whole training process ends at 1,300,000 iterations. For the dataset of Places365-standard, we use step size of 150,000 to decrease the learning rate and the whole train process stops at 600,000 iterations. To speed the training process, we use the multi-GPU extension~\cite{WangX16} of Caffe~\cite{JiaSDKLGGD14} toolbox for our CNN training~\footnote{\url{https://github.com/yjxiong/caffe}}. For testing our models, we use the common 5 crops (4 corners and 1 center) and their horizontal flipping for each image at a single scale, thus having 10 crops in total for each image. The final score is obtained by taking average over the predictions of 10 crops.

\subsection{Evaluation on multi-resolution CNNs}

{\bf Two-resolution CNNs.} We begin our experimental study by investigating the effectiveness of two-resolution CNNs on the validation sets of the ImageNet-1k, Places205, Places401, and Places365. Specifically, we study three architectures on all these datasets: (1) normal BN-Inception, which is trained from $256 \times 256$ images, (2) deeper BN-Inception, which has a deeper structure and is trained from $384 \times 384$ images, and (3) two-resolution CNN, which is the combination of both models, by using equal fusion weights. 

The results are summarized in Table \ref{tbl:mrcnn}. First, from comparison of normal BN-Inception and deeper BN-Inception, we conclude that CNNs trained from fine resolution images ($384 \times 384$) are able to yield better performance than those trained by coarse resolution images ($256 \times 256$) on all three datasets. Such superior performance may be ascribed to the fact that fine resolution images contain  richer information of visual content and more meaningful local details. In addition, the deeper BN-Inception is able to exhibit higher modeling capacity by using a deeper model, making it more powerful to capture complicated scene content. Second, we take an arithmetic average over the scores of normal BN-Inception and deeper BN-Inception as the results of two-resolution CNNs. This simple fusion scheme further boosts the recognition performance on three datasets. These improvements indicate that the multi-level information captured by two CNNs trained from different resolution images are strongly complementary to each other. Finally, we further compare our two-resolution CNNs with other baselines (such as AlexNet and VGGNet-16) on three datasets, and our approach outperforms these baselines by a large margin. It is worth noting that our two-resolution CNN is a modular learning framework that is readily applicable to any existing network structure to enhance its capacity.

{\bf Multi-resolution CNNs.} After verifying effectiveness of training CNNs from two resolutions on four datasets, we perform an extensive study to investigate the performance of training CNNs from multiple resolutions (128, 256, 384, 512) on the datasets of ImageNet-1k (object centric) and Places365 (scene centric). To keep the setup simple and comparison fair, the network architectures of different resolutions are built based on the original BN-Inception structure~\cite{IoffeS15} by stacking several convolutional layers after the Inception5b layer for the fine resolution CNNs (i.e., $384 \times 384$ and $512 \times 512$), and changing the pooling size of global pooling layer into $3 \times 3$ for the coarse resolution CNNs (i.e., $128 \times 128$). The experimental results are summarized in Table~\ref{tbl:mrcnn1}. From the results, we see that CNNs trained at a finer resolution can yield a better performance for both object centric and scene centric datasets. Meanwhile, we also notice that the resolution of $512 \times 512$ is able to improve recognition accuracy on the dataset of Places365, while the top5 classification accuracy already saturates on the dataset of ImageNet-1k. We also combine the recognition results from four resolutions and is able to obtain better performance than two-resolution CNNs. Our empirical study highlights the importance of image resolution in the network design and may provide some hints for the future work on image recognition with deep learning.

\subsection{Evaluation on knowledge guided disambiguation}

\begin{table}[t]
\begin{center}
\caption{Top5 classification error of different knowledge guided disambiguation techniques on the dataset of Places401.}
\label{tbl:kd}
\begin{tabular}{|l|c|}
  \hline
  Method & Places401 Validation  \\
  \hline
  \hline
  (A0) Normal BN-Inception ($256 \times 256$)  & 17.4\% \\
  \hline \hline
  (A1) Normal BN-Inception + object networks & 17.4\% \\
  (A2) Normal BN-Inception + scene networks & 16.7\% \\
  (A3) Normal BN-Inception + confusion matrix & 17.3\% \\
  \hline
  \hline
  Fusion of (A0) and (A1) & 16.7\% \\
  Fusion of (A0) and (A2) & 16.3\% \\
  Fusion of (A0) and (A3) & 16.6\% \\
  \hline \hline
  (B0) Deeper BN-Inception ($384 \times 384$) &  16.7\% \\
  \hline \hline
  (B1) Deeper BN-Inception + object networks & 16.3\% \\
  (B2) Deeper BN-Inception + scene networks & 16.1\% \\
  \hline
  \hline
  Fusion of (B0) and (B1) & 15.9\% \\
  Fusion of (B0) and (B2) & 15.8\% \\
  \hline
  \hline
  Fusion of (A0) and (B0) & 16.3\% \\
  Fusion of (A1) and (B1) & 16.1\% \\
  Fusion of (A2) and (B2) & 15.7\% \\
  \hline
\end{tabular}
\end{center}
\end{table}

\begin{table}[t]
\begin{center}
\caption{Top5 classification error of different knowledge guided disambiguation techniques on the dataset of Places365.}
\label{tbl:kd1}
\begin{tabular}{|l|c|}
  \hline
  Method & Places365 Validation  \\
  \hline
  \hline
  (A0) Normal BN-Inception ($256 \times 256$)  & 14.3\% \\
  \hline \hline
  (A1) Normal BN-Inception + object networks & 14.1\% \\
  (A2) Normal BN-Inception + scene networks & 13.4\% \\
  \hline
  \hline
  Fusion of (A0) and (A1) & 13.4\% \\
  Fusion of (A0) and (A2) & 12.9\% \\
  \hline \hline
  (B0) Deeper BN-Inception ($384 \times 384$) &  14.0\% \\
  \hline \hline
  (B1) Deeper BN-Inception + object networks & 13.6\% \\
  (B2) Deeper BN-Inception + scene networks & 13.0\% \\
  \hline
  \hline
  Fusion of (B0) and (B1) & 13.0\% \\
  Fusion of (B0) and (B2) & 12.8\% \\
  \hline
  \hline
  Fusion of (A0) and (B0) & 13.2\% \\
  Fusion of (A1) and (B1) & 13.0\% \\
  Fusion of (A2) and (B2) & 12.4\% \\
  \hline
\end{tabular}
\end{center}
\end{table}

\begin{figure*}[t]
\centering
\includegraphics[width=0.33\linewidth]{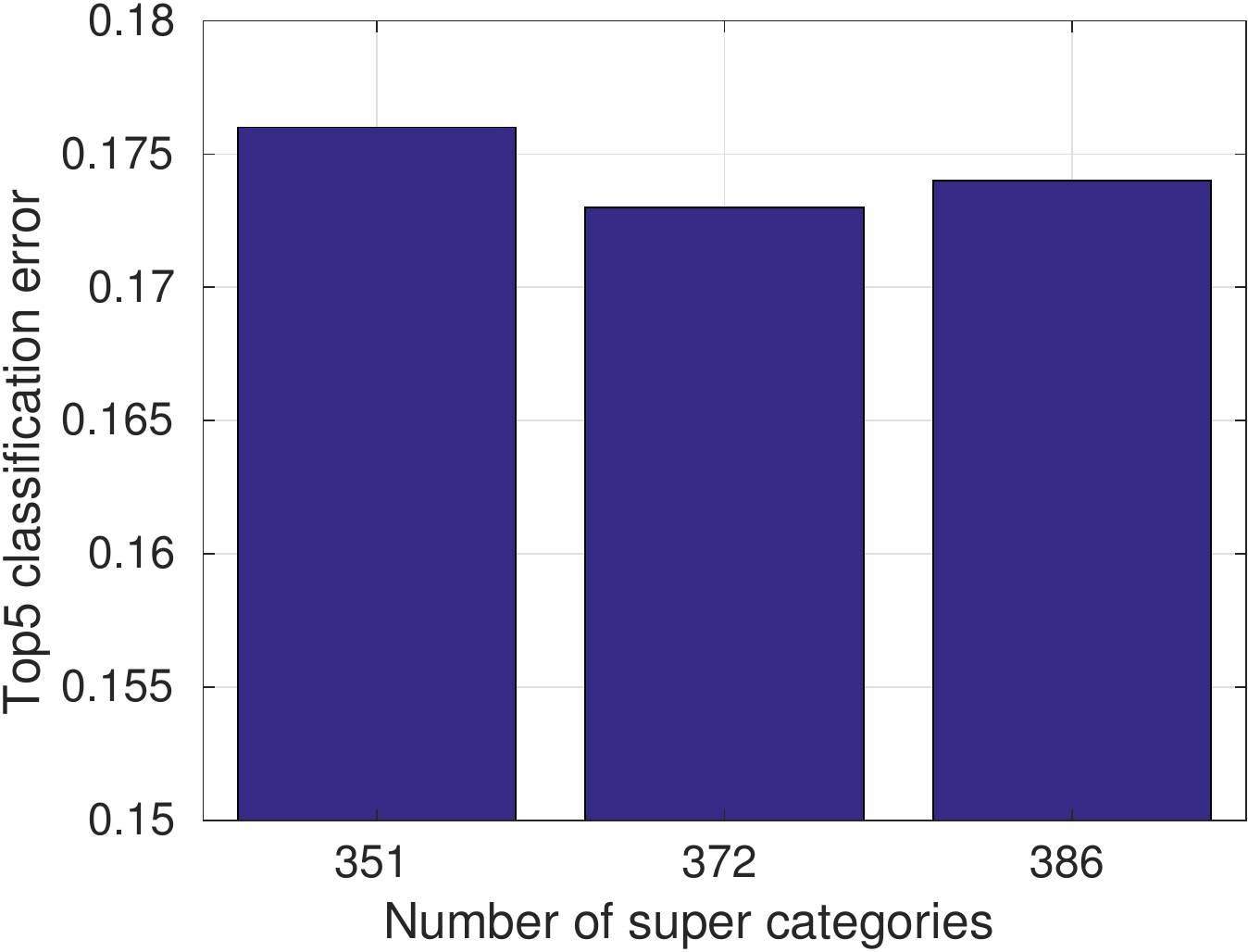}
\includegraphics[width=0.325\linewidth]{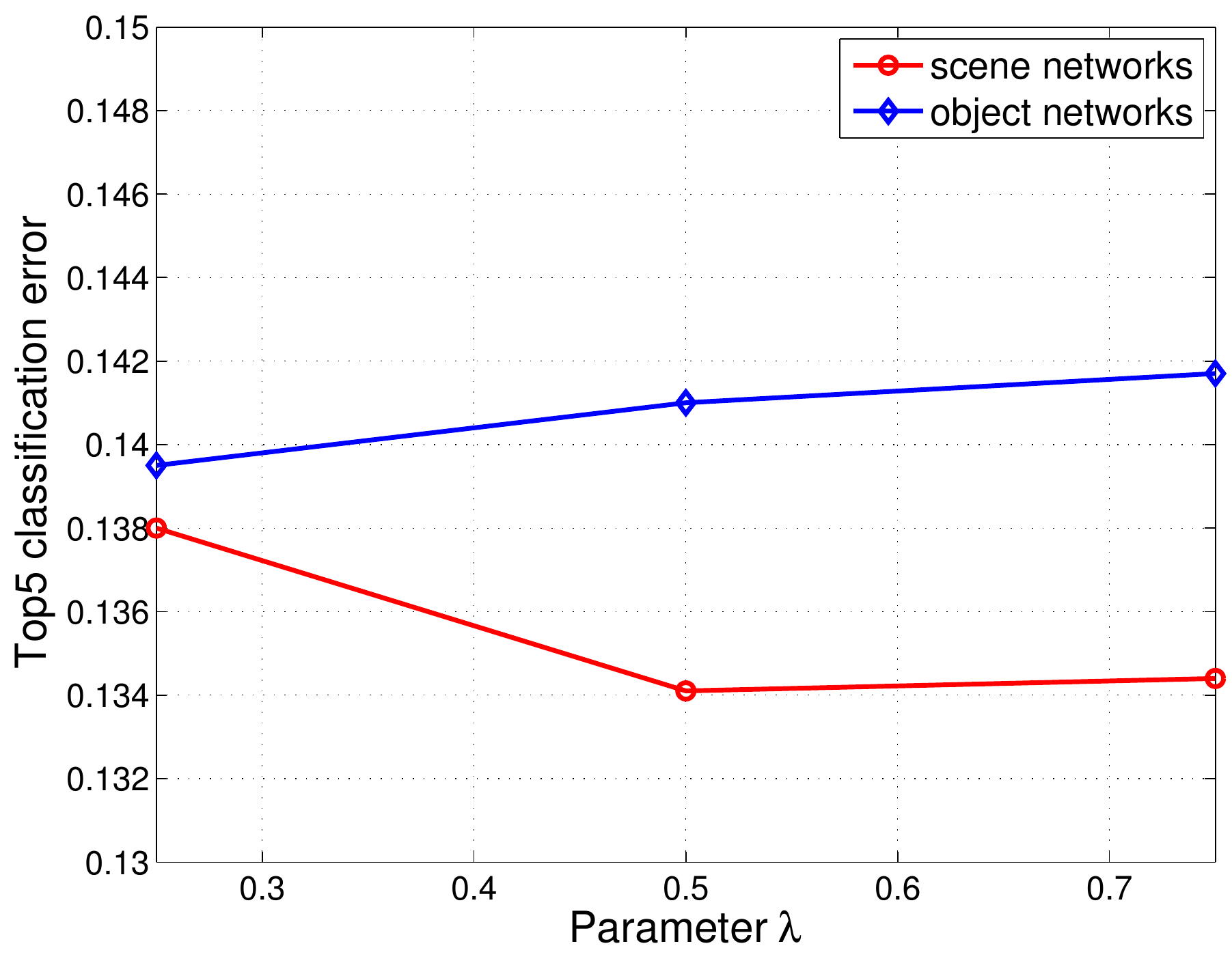}
\includegraphics[width=0.325\linewidth]{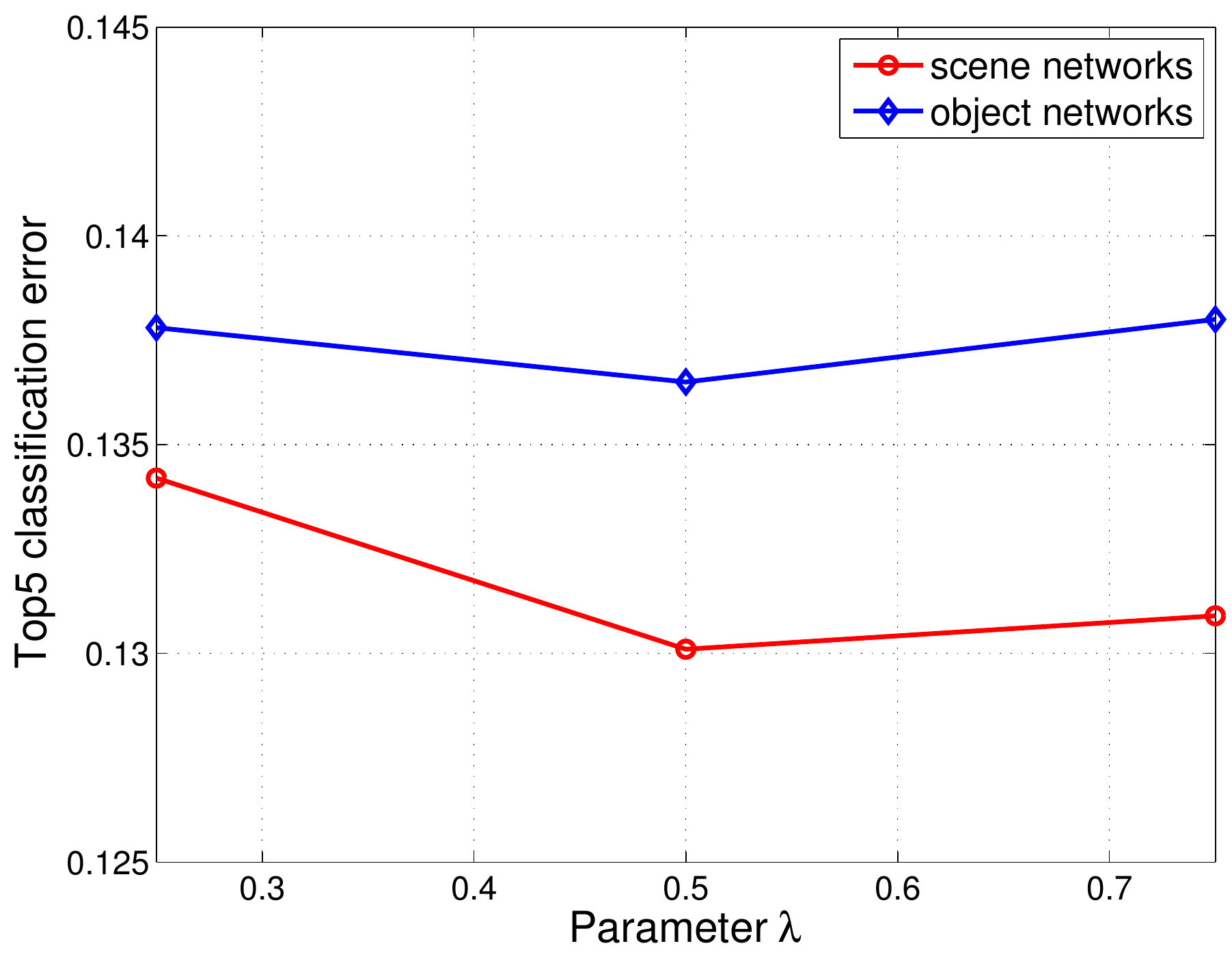}
\caption{{\bf Exploration study.} {\em Left:} Top5 classification error of different numbers of super categories on the Places401 dataset; {\em Center:} Top5 classification error of extra network based disambiguation methods at resolution of $256 \times 256$ on the Places365 dataset; {\em Right:} Top5 classification error of extra network based disambiguation methods at resolution of $384 \times 384$ on the Places365 dataset. }
\label{fig:exploration}
\end{figure*}

We now turn to study the effectiveness of our proposed knowledge guided disambiguation techniques described in Section \ref{sec:kd}. To handle the issue of category ambiguity in large-scale scene recognition, we proposed two disambiguation techniques, one of which is based on the knowledge of confusion matrix on the validation dataset, and the other one explores knowledge from extra networks. As the label ambiguity is particularly important for large-scale scene recognition, we perform experiment on the Places401 and Places365 dataset.

{\bf Knowledge from confusion matrix.} 
We first verify the effectiveness of merging similar categories into super categories on the dataset of Places401. According to the confusion matrix, we merge 401 scene categories into 386, 372, and 351 super categories by setting the threshold $\tau$ in Algorithm~\ref{alg:merge} to 0.3, 0.5, and 0.7. The results are shown in the left of Figure~\ref{fig:exploration}. We see that containing 372 super categories achieves the lowest top5 classification error (17.3\%) and properly setting the parameter of threshold $\tau$ is crucial to improve recognition accuracy. Therefore, in the remaining experiment, we fix the parameter $\tau$ as 0.5. 

We fix the number of super categories as 372 (A3) and compare with the original network (A0) as shown in Table~\ref{tbl:kd}. As can be found, by utilizing knowledge from confusion matrix, the performance of normal BN-Inception network is improved slightly. This result is a little bit surprising, as we use less category information, but still obtain higher performance. This result indicates that label ambiguity may leads to the problem of over-fitting on subtle differences in an effort to distinguish visually closed categories (e.g., baseball field vs. stadium baseball). But these fine-scale differences may not generalize well on unseen images, so as to decrease the recognition performance on testing set. This agrees with the findings of the Places team and their latest version (Places365) has already merged very similar scene categories. 

{\bf Knowledge from extra networks.} 
In our second disambiguation approach, we utilize two extra networks: one pre-trained on the ImageNet-1k dataset (referred as object network) and one pre-trained on the Places205 (referred as scene network). We use the outputs of these extra networks as soft labels to guide the training of our CNNs in a multi-task learning framework. An important parameter in this framework is $\lambda$ in Equation~(\ref{equ:mt}). We {\em first} perform exploration study to determine this parameter on the dataset of Places365. The experimental results are reported in Figure~\ref{fig:exploration}. We can see that the top5 classification error of disambiguation by object network is less sensitive to the parameter $\lambda$ compared with scene network. The parameter of $\lambda=0.5$ is the best choice for scene network disambiguation for both normal BN-Inception and deeper BN-Inception architectures. Therefore, we fix the parameter $\lambda$ as 0.5 in the remaining experimental study.

{\em Next}, we give a detailed analysis about the disambiguation techniques on the datasets of Places401 and Places365. The numerical results are summarized in Table~\ref{tbl:kd} and Table~\ref{tbl:kd1}. From these result analysis, several conclusions can be drawn as follows:
\begin{itemize}
\item First, our knowledge network based disambiguation techniques are able to improve the recognition accuracy of original networks on both datasets of Places401 and Places365. Although the new Places365 dataset already merges very similar categories, our disambiguation technique is still capable of regularizing the CNN training and improving the generalization performance on this dataset.
\item Second, comparing the performance of disambiguation with different knowledge networks, we see that scene network can yield better performance than object network. For instance, on the dataset of Places365, the network of A2 obtains the top5 classification error of 13.4\%, while the error rate of A1 is 14.1\%. This may be ascribed to the fact that the scene classes from the Places are more correlated with the categories in the Places2 than those object classes in the ImageNet.
\item Finally, we explore different network architectures, including normal BN-Inception ($256 \times 256$), deeper BN-Inception ($384 \times 384$), and two-resolution CNNs. We find our proposed knowledge disambiguation method is able to improve the performance for all network architectures. For example, on the dataset of Places365, the fusion of A2 and B2 achieves the classification error of 12.4\%, which is lower than the error rate 13.1\% of fusion of A0 and B0.
\end{itemize}

{\em Finally}, to fully unleash the benefits of our knowledge disambiguation method, we perform model fusion between normally trained CNNs and knowledge guided CNNs. From these results, we see that those knowledge guided CNNs are complementary to those normally trained CNNs and the fusion of them can improve performance considerably. For the normal BN-Inception architecture, the best combination of (A0) and (A2) reduces the top5 error rate from 17.4\%  to 16.3\% on the dataset Places401 and from 14.3\% to 12.9\% on the dataset of Places365. With the deeper BN-Inception network, on the dataset of Places401, the best combination of (B0) and (B2) achieves a top5 error of 15.8\%, compared to the original 16.7\%, and on the dataset of Places365, similar improvement is achieved as well (12.8\% vs. 14.0\%). These excellent fusion results suggest that our proposed knowledge guided disambiguation techniques not only improve the performance of the original models, but also provide reliable complementary models that build stronger model ensembles.

\subsection{Results at the Places2 challenge 2015}

\begin{table}[t]
\begin{center}
\caption{Classification error of different teams at the Places2 challenge 2015.}
\label{tbl:challenge}
\resizebox{\linewidth}{!}{
\begin{tabular}{|l|l|c|c|}
  \hline
  Rank & Team & Places2 Test & Places2 Val \\
  \hline
  1 & WM \cite{ShenLH15} & {\bf 16.9\%} & 15.7\% \\
  \hline
  2 & SIAT\_MMLAB (A0+A1+A2+A3+B0) & 17.4\% & 15.8\%  \\
  - & Post submission (B0+B1+B2) & - & {\bf 15.5\%} \\
  \hline
  3 & Qualcomm & 17.6\% & - \\
  \hline
  4 & Trimps-Soushen & 18.0\% & - \\
  \hline
  5 & NTU-Rose & 19.3\% & - \\
  \hline
\end{tabular}
}
\end{center}
\end{table}

To provide more convincing results, we investigate the performance of our whole pipeline, including both the multi-resolution CNNs and knowledge guided disambiguation techniques, on a large-scale scene recognition challenge. Here we report our results on the Places2 challenge 2015, which is the largest scene recognition challenge, and was held in conjunction with the ImageNet large-scale visual recognition challenge (ILSVRC)~\cite{ILSVRC15}.

Results of the Places2 challenge 2015 are summarized in Table \ref{tbl:challenge}. Our SIAT\_MMLAB team secured the second place and our challenge solution corresponds to the combination of models A0+A1+A2+A3+B0. Our solution was outperformed by the winner method~\cite{ShenLH15}, with a 0.5\% gap in top5 error in test phase. The winner method exploited a multi-scale cropping strategy which leads to large performance gains, while we just simply used a single-scale cropping method in all our experiments.

In addition, it is worth noting that our submission did not contain our best model architecture of B2, due to deadline of the challenge. After the challenge, we finished the training of B2 model, which achieves better performance on the validation dataset. Finally, we achieve the performance of 15.5\% top5 error on the validation set by using the model fusion of B0+B1+B2, surpassing the best result of the winner method (15.7\%).

\subsection{Results at LSUN challenge 2016}

\begin{table}[t]
\begin{center}
\caption{Classification accuracy of different pre-trained models on the validation set of LSUN classification dataset.}
\label{tbl:lsun1}
\begin{tabular}{|l|c|}
  \hline
  Pre-trained Model & Top1 Accuracy\\
  \hline \hline
  (A0) Normal BN-Inception ($256 \times 256$)  & 89.9\% \\
  (A1) Normal BN-Inception + object networks & 90.1\% \\
  (A2) Normal BN-Inception + scene networks & 90.4\% \\
  \hline \hline
  (B0) Deeper BN-Inception ($384 \times 384$) &  90.5\% \\
  (B1) Deeper BN-Inception + object networks & 90.7\% \\
  (B2) Deeper BN-Inception + scene networks & 90.9\% \\
  \hline \hline
  (A0+B0) & 91.0\% \\
  Fusion all & 91.8\% \\
  \hline
\end{tabular}
\end{center}
\end{table}

In this subsection we further present our results on another important scene recognition challenge, namely Large-Scale Scene Understanding (LSUN) challenge, which aims to provide a different benchmark for large-scale scene classification and understanding~\footnote{\url{http://lsun.cs.princeton.edu}}. The LSUN classification dataset \cite{YuZSSX15} contains 10 scene categories, such as dining room, bedroom, chicken, outdoor church, and so on. For training data, each category contains a huge number of images, ranging from around 120,000 to 3,000,000, which is significantly unbalanced. The validation data includes 300 images, and the test data has 1000 images for each category. The evaluation of LSUN classification challenge is based on top1 classification accuracy. 

In order to verify the effectiveness of our proposed multi-resolution CNN and knowledge guided disambiguation strategy, we transfer the learned representations on the Places401 dataset to the classification task of the LSUN challenge. Specifically, to reduce computational cost and balance the training samples from each category, we randomly sample 100,000 images from each category as our training data. Then, we use our learned CNNs on the Places401 dataset as pre-training models, and fine tune them on the LSUN dataset. The learning rate is initialized as 0.1, which is decreased by $\frac{1}{10}$ every 60,000 iterations. The batch size is set as 256. The whole training process stops at 180,000 iterations. During the test phase, by following previous common cropping techniques, we crop 5 regions with their horizontal flipping, and use 3 different scales for each image. We take an average over these prediction scores of different crops as the final result of an input image.

We report the performance of our fine-tuned CNNs on the validation set of LSUN dataset, building on various Places401 pre-trained models. Results are presented in Table \ref{tbl:lsun1}. First, by comparing the performance of CNNs at different resolutions, we find that the deeper BN-Inception networks learned on finer resolution images yield better results than the normal BN-Inception networks (89.9\% vs. 90.5\%). Second, considering the strategy of knowledge guided disambiguation, both object and scene guided CNNs are capable of bringing improvements (around 0.5\%) over those non-guided CNNs. Finally, we fuse prediction scores of multiple networks, and obtain the final performance with a top1 accuracy of 91.8\% on the validation set of LSUN dataset. 

\begin{table}[t]
\begin{center}
\caption{Classification accuracy of different teams at the LSUN challenge 2016.}
\label{tbl:lsun2}
\begin{tabular}{|l|c|c|c|}
  \hline
  Rank & Team & Year & Top1 Accuracy\\
  \hline
  1 & SIAT\_MMLAB & 2016 & {\bf 91.6\%} \\
  2 & SJTU-ReadSense & 2016 & 90.4\% \\
  3 & TEG Rangers & 2016 & 88.7\% \\
  4 & ds-cube & 2016 & 83.0\% \\
  \hline
  1 & Google & 2015 & 91.2\% \\
  \hline 
\end{tabular}
\end{center}
\end{table}

We further provide the results of our method on the test set of LSUN dataset, by fusing all our models. We compare our result against those of other teams attending this challenge in Table \ref{tbl:lsun2}. Our SIAT\_MMLAB team obtained the performance of 91.6\% top1 accuracy which secures the 1$^{st}$ place at this challenge. This excellent result strongly demonstrates the effectiveness of our proposed solution for large-scale scene recognition. Furthermore, we obtain an improvement of 0.4\% top1 accuracy (evaluated on a same database) over the strong baseline achieved by Google team, who was the winner of last LSUN challenge in 2015. Our advantages are built on the proposed multi-resolution structure and knowledge guided disambiguation strategy, by using a similar Inception architecture.

\subsection{Generalization analysis}

\begin{table}[t]
\begin{center}
\caption{Comparison of the transferred representations of our model with other methods on the MIT67 and SUN397 datasets.}
\label{tbl:other}
\begin{tabular}{|l|c|c|}
  \hline
  Model & MIT Indoor67 & ~~~SUN397~~~ \\
  \hline
  ImageNet-VGGNet-16 \cite{SimonyanZ14a} ~~~~~& 67.7\% & 51.7\% \\
  Places205-AlexNet \cite{ZhouLXTO14} & 68.2\% & 54.3\% \\
  Places205-GoogLeNet \cite{guo2016locally}  & 74.0\% & 58.8\% \\
  Places205-CNDS-8 \cite{WangLTL15a} & 76.1\% & 60.7\% \\
  Places205-VGGNet-16 \cite{WangGH015} & 81.2\% &  66.9\% \\
  Places365-VGGNet-16 \cite{Places2} & 76.5\% & 63.2\% \\
  Hybrid1365-VGGNet-16 \cite{Places2} & 77.6\% & 61.7\% \\
  DAG-VGGNet19 \cite{YangR15} & 77.5\% & 56.2\% \\
  MS-DSP \cite{gao2015deep} & 78.3\% & 59.8\% \\
  LS-DHM \cite{guo2016locally} & 83.8\% & 67.6\% \\
  VSAD~\cite{WangWWZ016} & 84.9\% & 71.7\% \\
  Multiple Models \cite{XieZYL16} & 86.0\% & 70.7\% \\
  Three \cite{Herranz_2016_CVPR} & 86.0\% & 70.2\% \\
  \hline
  Places365-Deeper-BN-Inception (B2) & 84.8\% & 71.7\% \\
  Places401-Deeper-BN-Inception (B2) & {\bf 86.7\%} & {\bf 72.0\%} \\
  \hline
\end{tabular}
\end{center}
\end{table}

Extensive experimental results have demonstrated the effectiveness of our proposed method on large-scale datasets, by eigher training from scratch (using the Places2), or adaption with fine tuning (on the LSUN). In this subsection, we evaluate the generalization ability of our learned models, by directly applying them on two other databases: the MIT Indoor67 \cite{QuattoniT09} and SUN397~\cite{XiaoHEOT10},
which have been used as standard benchmarks for scene recognition for many years.  Most recent methods reported and compared their results on these two datasets. 

The MIT Indoor67 \cite{QuattoniT09} contains 67 indoor-scene categories and has a total of 15,620 images, with at least 100 images per category. Following the original evaluation protocol, we use 80 images from each category for training, and another 20 images for testing.  The SUN397 \cite{XiaoHEOT10} has a large number of scene categories by including 397 categories and totally 108,754 images. Each category has at least 100 images. We follow the standard evaluation protocol provided in the original paper by using 50 training and 50 test images for each category. The partitions are fixed and publicly available from the original paper~\cite{XiaoHEOT10}. Finally, the average classification accuracy of ten different tests is reported.

In this experiment, we directly use the trained B2 models on the datasets of Places365 and Places401 as generic feature extractors, without fine tuning them on the target dataset. Specifically, the test images are first resized as $384 \times 384$. We then crop image regions of different scales ($384 \times 384$, $346 \times 346$, and $336 \times 336$) from the input images. After this, these image regions are resized as $336 \times 336$ and fed into our pre-trained CNNs for feature extraction. We utilize the activation of global pooling as global representation. These representations of different regions are averaged and normalized with $\ell_2$-norm, which is used as final representation of the input image. For classifier, we use the linear SVM with LIBSVM implementation \cite{ChangL11}.

The experimental results are summarized in Table \ref{tbl:other}. We compare the transfered representations of our model trained on the Places401 and Places365 datasets against other deep models (e.g., VGGNet \cite{SimonyanZ14a} and GoogLeNet \cite{SzegedyLJSRAEVR15}) trained on various datasets (e.g.,  the Places or ImageNet). As shown in Table \ref{tbl:other}, our transferred model achieves best performance among all methods, demonstrating that our method generalizes better than the others. To the best of our knowledge, the performance of 86.7\% on the MIT Indoor67 and 72.0\% on the SUN397 are the best results on both datasets, which advance the state of the art substantially. We believe that our excellent performance is valuable to scene recognition community, and future large-scale recognition algorithm can be built on our pre-trained models.

\subsection{Failure case analysis}

\begin{figure*}[t]
\includegraphics[width=\textwidth]{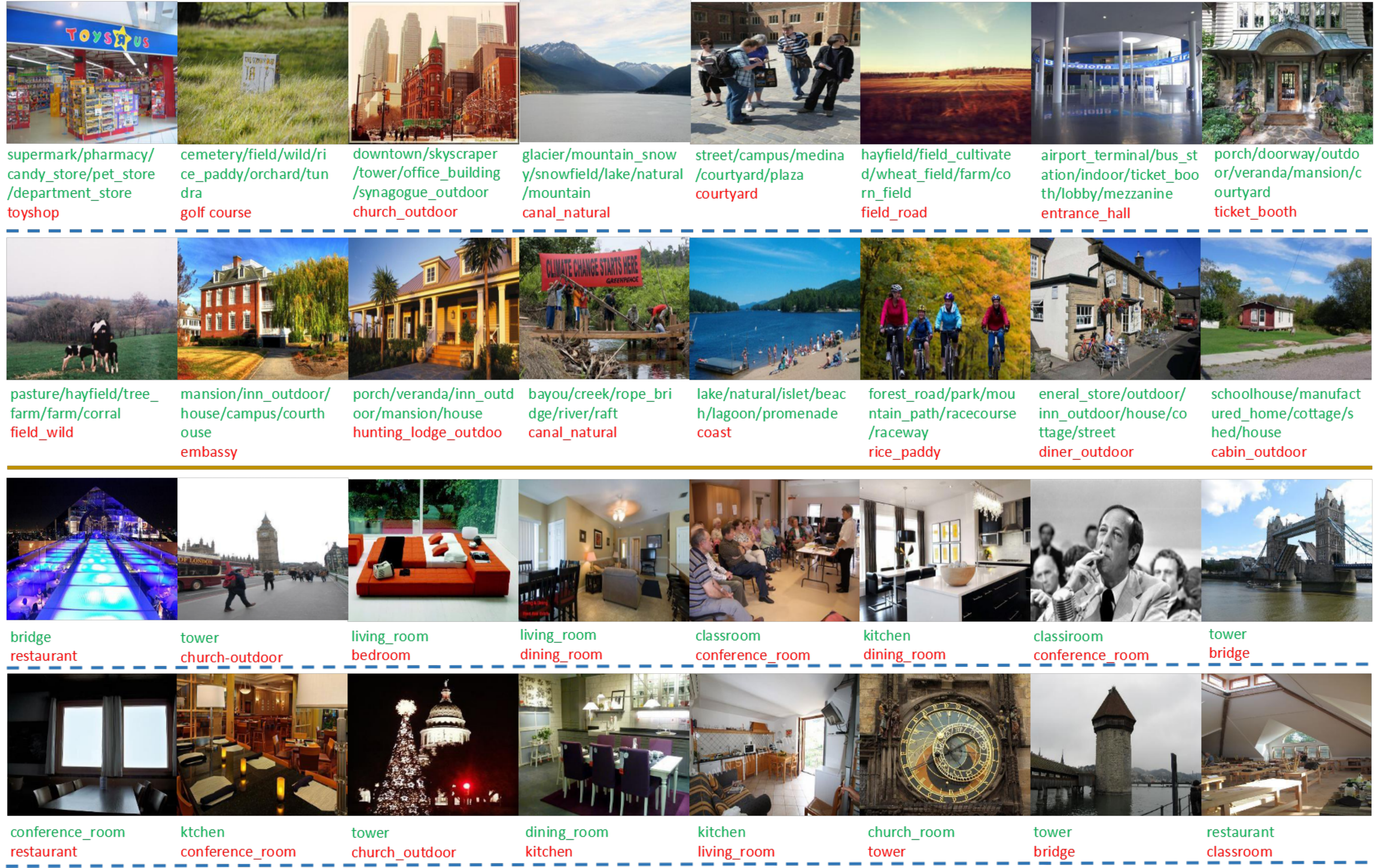}
\caption{Examples of images that our method fail to predict the correct labels within 5 guesses. \textbf{Top rows}: we show 16 failure cases (under {\bf top5} evaluation) on the validation set of the Places401 dataset. The predicted labels (in green) are sorted according to their confidence score and the correct label is labeled in red. \textbf{Bottom rows}: we give 16 examples that our method fail to predict the correct labels (under {\bf top1} evaluation) on the validation set of the LSUN dataset. The predicted label is marked with green color, while the ground truth is with red color.}
\label{fig:ex_error}
\end{figure*}

Finally, we present a number of failure examples by our method from the datasets of Places401 and LSUN. These examples are illustrated in Figure \ref{fig:ex_error}. From these examples, we notice that some scene classes are easily confused with others. In the Places401 database, the categories of \textit{supermarket},\textit{ pet-store}, \textit{toyshop} look very similar from outdoor appearance. The classes of {\em downtown}, {\em building}, and {\em skyscraper} may co-occur in many images. 
Thus, scene image often contains complicated visual content which is difficult to be described clearly by a single category label, and multi-label classification can be applied to ambiguous categories. For the dataset of LSUN, the classes of {\em bridge} and {\em tower} are highly ambiguous in some cases. Similarly, the category of {\em conference room} is sometimes confused with the {\em classroom} category, due to their closed spatial layout and common objects contained. Overall, from these failure cases, we can see that scene recognition is still a challenging problem, and label ambiguity is a crucial issue in large-scale scene recognition. Meanwhile, scene recognition sometimes is essentially a kind of multi-label classification problem and in the future we may consider multi-label classification framework~\cite{YangZC16,YangZZGWC15} for scene recognition.

\section{Conclusions}
\label{sec:con}

In this paper we have studied the problem of scene recognition on large-scale datasets such as the Places, Places2, and LSUN. Large-scale scene recognition suffers from two major problems: visual inconsistence (large intra-class variation) and label ambiguity (small inter-class variation). We developed powerful multi-resolution knowledge guided disambiguation framework that effectively tackle these two crucial issues. We introduced multi-resolution CNNs which are able to capture visual information from different scales. Furthermore, we proposed two knowledge guided disambiguation approaches to exploit extra knowledge, which guide CNNs training toward a better optimization,  with improved generalization ability. 

We conducted extensive experiments on three large-scale scene databases: the Places2, Places, and LSUN, and directly transferred our learned representation to two widely-used standard scene benchmarks: the MIT Indoor67 and SUN397.  Our method achieved superior performance on all five benchmarks, advancing the state-of-the-art results substantially. These results convincingly demonstrate the effectiveness of our method. Importantly, our method attended two most domain-influential challenges for large-scale scene recognition. We  achieved the second place at the Places2 challenge in ILSVRC 2015, and the first place at the LSUN challenge in CVPR 2016. These impressive results further confirm the strong capability of our method.


%



\ifCLASSOPTIONcaptionsoff
  \newpage
\fi



%

\bibliographystyle{IEEEtran}
\bibliography{references}

\begin{thebibliography}{10}
\providecommand{\url}[1]{#1}
\csname url@samestyle\endcsname
\providecommand{\newblock}{\relax}
\providecommand{\bibinfo}[2]{#2}
\providecommand{\BIBentrySTDinterwordspacing}{\spaceskip=0pt\relax}
\providecommand{\BIBentryALTinterwordstretchfactor}{4}
\providecommand{\BIBentryALTinterwordspacing}{\spaceskip=\fontdimen2\font plus
\BIBentryALTinterwordstretchfactor\fontdimen3\font minus
  \fontdimen4\font\relax}
\providecommand{\BIBforeignlanguage}[2]{{%
\expandafter\ifx\csname l@#1\endcsname\relax
\typeout{** WARNING: IEEEtran.bst: No hyphenation pattern has been}%
\typeout{** loaded for the language `#1'. Using the pattern for}%
\typeout{** the default language instead.}%
\else
\language=\csname l@#1\endcsname
\fi
#2}}
\providecommand{\BIBdecl}{\relax}
\BIBdecl

\bibitem{LiPT05}
F.~Li and P.~Perona, ``A bayesian hierarchical model for learning natural scene
  categories,'' in \emph{CVPR}, 2005, pp. 524--531.

\bibitem{WuR11}
J.~Wu and J.~M. Rehg, ``{CENTRIST:} {A} visual descriptor for scene
  categorization,'' \emph{{IEEE} Trans. Pattern Anal. Mach. Intell.}, vol.~33,
  no.~8, pp. 1489--1501, 2011.

\bibitem{ZhangZS14}
L.~Zhang, X.~Zhen, and L.~Shao, ``Learning object-to-class kernels for scene
  classification,'' \emph{{IEEE} Trans. Image Processing}, vol.~23, no.~8, pp.
  3241--3253, 2014.

\bibitem{ZuoWSZYJ14}
Z.~Zuo, G.~Wang, B.~Shuai, L.~Zhao, Q.~Yang, and X.~Jiang, ``Learning
  discriminative and shareable features for scene classification,'' in
  \emph{ECCV}, 2014, pp. 552--568.

\bibitem{GongWGL14}
Y.~Gong, L.~Wang, R.~Guo, and S.~Lazebnik, ``Multi-scale orderless pooling of
  deep convolutional activation features,'' in \emph{ECCV}, 2014, pp. 392--407.

\bibitem{XieWGZT14}
L.~Xie, J.~Wang, B.~Guo, B.~Zhang, and Q.~Tian, ``Orientational pyramid
  matching for recognizing indoor scenes,'' in \emph{CVPR}, 2014, pp.
  3734--3741.

\bibitem{YangR15}
S.~Yang and D.~Ramanan, ``Multi-scale recognition with {DAG-CNNs},'' in
  \emph{ICCV}, 2015, pp. 1215--1223.

\bibitem{ShenLH15}
L.~Shen, Z.~Lin, and Q.~Huang, ``Relay backpropagation for effective learning
  of deep convolutional neural networks,'' in \emph{ECCV}, 2016, pp. 467--482.

\bibitem{ZuoSWLWWC16}
Z.~Zuo, B.~Shuai, G.~Wang, X.~Liu, X.~Wang, B.~Wang, and Y.~Chen, ``Learning
  contextual dependence with convolutional hierarchical recurrent neural
  networks,'' \emph{{IEEE} Trans. Image Processing}, vol.~25, no.~7, pp.
  2983--2996, 2016.

\bibitem{guo2016locally}
S.~Guo, W.~Huang, L.~Wang, and Y.~Qiao, ``Locally supervised deep hybrid model
  for scene recognition,'' \emph{{IEEE} Trans. Image Processing}, vol.~26,
  no.~2, pp. 808--820, 2017.

\bibitem{WangWWZ016}
Z.~Wang, L.~Wang, Y.~Wang, B.~Zhang, and Y.~Qiao, ``Weakly supervised
  patchnets: Describing and aggregating local patches for scene recognition,''
  \emph{CoRR}, vol. abs/1609.00153, 2016.

\bibitem{OlivaT01}
A.~Oliva and A.~Torralba, ``Modeling the shape of the scene: {A} holistic
  representation of the spatial envelope,'' \emph{International Journal of
  Computer Vision}, vol.~42, no.~3, pp. 145--175, 2001.

\bibitem{Torralba03}
A.~Torralba, ``Contextual priming for object detection,'' \emph{International
  Journal of Computer Vision}, vol.~53, no.~2, pp. 169--191, 2003.

\bibitem{FelzenszwalbGMR10}
P.~F. Felzenszwalb, R.~B. Girshick, D.~A. McAllester, and D.~Ramanan, ``Object
  detection with discriminatively trained part-based models,'' \emph{{IEEE}
  Trans. Pattern Anal. Mach. Intell.}, vol.~32, no.~9, pp. 1627--1645, 2010.

\bibitem{WangWLC11}
L.~Wang, Y.~Wu, T.~Lu, and K.~Chen, ``Multiclass object detection by combining
  local appearances and context,'' in \emph{ACM Multimedia}, 2011, pp.
  1161--1164.

\bibitem{wang2015object}
L.~Wang, Z.~Wang, W.~Du, and Y.~Qiao, ``Object-scene convolutional neural
  networks for event recognition in images,'' in \emph{CVPR Workshops}, 2015,
  pp. 30--35.

\bibitem{XiongZLT15}
Y.~Xiong, K.~Zhu, D.~Lin, and X.~Tang, ``Recognize complex events from static
  images by fusing deep channels,'' in \emph{CVPR}, 2015, pp. 1600--1609.

\bibitem{WangWQG16}
L.~Wang, Z.~Wang, Y.~Qiao, and L.~V. Gool, ``Transferring object-scene
  convolutional neural networks for event recognition in still images,''
  \emph{CoRR}, vol. abs/1609.00162, 2016.

\bibitem{Wang0T16}
L.~Wang, Y.~Qiao, and X.~Tang, ``{MoFAP}: {A} multi-level representation for
  action recognition,'' \emph{International Journal of Computer Vision}, vol.
  119, no.~3, pp. 254--271, 2016.

\bibitem{ZhangWCDL16}
Y.~Zhang, L.~Cheng, J.~Wu, J.~Cai, M.~N. Do, and J.~Lu, ``Action recognition in
  still images with minimum annotation efforts,'' \emph{{IEEE} Trans. Image
  Processing}, vol.~25, no.~11, pp. 5479--5490, 2016.

\bibitem{WangQT14}
L.~Wang, Y.~Qiao, and X.~Tang, ``Latent hierarchical model of temporal
  structure for complex activity classification,'' \emph{{IEEE} Trans. Image
  Processing}, vol.~23, no.~2, pp. 810--822, 2014.

\bibitem{ZhouLXTO14}
B.~Zhou, {\`{A}}.~Lapedriza, J.~Xiao, A.~Torralba, and A.~Oliva, ``Learning
  deep features for scene recognition using places database,'' in \emph{NIPS},
  2014, pp. 487--495.

\bibitem{Places2}
B.~Zhou, A.~Khosla, {\`{A}}.~Lapedriza, A.~Torralba, and A.~Oliva, ``Places: An
  image database for deep scene understanding,'' \emph{CoRR}, vol.
  abs/1610.02055, 2016.

\bibitem{lecun-98}
Y.~LeCun, L.~Bottou, Y.~Bengio, and P.~Haffner, ``Gradient-based learning
  applied to document recognition,'' \emph{Proceedings of the IEEE}, vol.~86,
  no.~11, pp. 2278--2324, 1998.

\bibitem{XiaoHEOT10}
J.~Xiao, J.~Hays, K.~A. Ehinger, A.~Oliva, and A.~Torralba, ``{SUN} database:
  Large-scale scene recognition from abbey to zoo,'' in \emph{CVPR}, 2010, pp.
  3485--3492.

\bibitem{IoffeS15}
S.~Ioffe and C.~Szegedy, ``Batch normalization: Accelerating deep network
  training by reducing internal covariate shift,'' in \emph{ICML}, 2015, pp.
  448--456.

\bibitem{DengDSLL009}
J.~Deng, W.~Dong, R.~Socher, L.~Li, K.~Li, and F.~Li, ``{ImageNet}: {A}
  large-scale hierarchical image database,'' in \emph{CVPR}, 2009, pp.
  248--255.

\bibitem{ILSVRC15}
O.~Russakovsky, J.~Deng, H.~Su, J.~Krause, S.~Satheesh, S.~Ma, Z.~Huang,
  A.~Karpathy, A.~Khosla, M.~Bernstein, A.~C. Berg, and L.~Fei-Fei, ``{ImageNet
  Large Scale Visual Recognition Challenge},'' \emph{International Journal of
  Computer Vision}, vol. 115, no.~3, pp. 211--252, 2015.

\bibitem{QuattoniT09}
A.~Quattoni and A.~Torralba, ``Recognizing indoor scenes,'' in \emph{CVPR},
  2009, pp. 413--420.

\bibitem{LazebnikSP06}
S.~Lazebnik, C.~Schmid, and J.~Ponce, ``Beyond bags of features: Spatial
  pyramid matching for recognizing natural scene categories,'' in \emph{CVPR},
  2006, pp. 2169--2178.

\bibitem{PariziOF12}
S.~N. Parizi, J.~G. Oberlin, and P.~F. Felzenszwalb, ``Reconfigurable models
  for scene recognition,'' in \emph{CVPR}, 2012, pp. 2775--2782.

\bibitem{PandeyL11}
M.~Pandey and S.~Lazebnik, ``Scene recognition and weakly supervised object
  localization with deformable part-based models,'' in \emph{ICCV}, 2011, pp.
  1307--1314.

\bibitem{SinghGE12}
S.~Singh, A.~Gupta, and A.~A. Efros, ``Unsupervised discovery of mid-level
  discriminative patches,'' in \emph{ECCV}, 2012, pp. 73--86.

\bibitem{JunejaVJZ13}
M.~Juneja, A.~Vedaldi, C.~V. Jawahar, and A.~Zisserman, ``Blocks that shout:
  Distinctive parts for scene classification,'' in \emph{CVPR}, 2013, pp.
  923--930.

\bibitem{KrizhevskySH12}
A.~Krizhevsky, I.~Sutskever, and G.~E. Hinton, ``{ImageNet} classification with
  deep convolutional neural networks,'' in \emph{NIPS}, 2012, pp. 1106--1114.

\bibitem{ZeilerF14}
M.~D. Zeiler and R.~Fergus, ``Visualizing and understanding convolutional
  networks,'' in \emph{ECCV}, 2014, pp. 818--833.

\bibitem{HeZR014}
K.~He, X.~Zhang, S.~Ren, and J.~Sun, ``Spatial pyramid pooling in deep
  convolutional networks for visual recognition,'' in \emph{ECCV}, 2014, pp.
  346--361.

\bibitem{SimonyanZ14a}
K.~Simonyan and A.~Zisserman, ``Very deep convolutional networks for
  large-scale image recognition,'' \emph{CoRR}, vol. abs/1409.1556, 2014.

\bibitem{SzegedyLJSRAEVR15}
C.~Szegedy, W.~Liu, Y.~Jia, P.~Sermanet, S.~Reed, D.~Anguelov, D.~Erhan,
  V.~Vanhoucke, and A.~Rabinovich, ``Going deeper with convolutions,'' in
  \emph{CVPR}, 2015, pp. 1--9.

\bibitem{HeZRS15a}
K.~He, X.~Zhang, S.~Ren, and J.~Sun, ``Delving deep into rectifiers: Surpassing
  human-level performance on imagenet classification,'' in \emph{ICCV}, 2015,
  pp. 1026--1034.

\bibitem{Szegedy_2016_CVPR}
C.~Szegedy, V.~Vanhoucke, S.~Ioffe, J.~Shlens, and Z.~Wojna, ``Rethinking the
  inception architecture for computer vision,'' in \emph{CVPR}, 2016, pp.
  2818--2826.

\bibitem{HeZRS15}
K.~He, X.~Zhang, S.~Ren, and J.~Sun, ``Deep residual learning for image
  recognition,'' in \emph{CVPR}, 2016, pp. 770--778.

\bibitem{ZhangWWCLND16}
Y.~Zhang, X.~Wei, J.~Wu, J.~Cai, J.~Lu, V.~A. Nguyen, and M.~N. Do, ``Weakly
  supervised fine-grained categorization with part-based image
  representation,'' \emph{{IEEE} Trans. Image Processing}, vol.~25, no.~4, pp.
  1713--1725, 2016.

\bibitem{HintonVD15}
G.~E. Hinton, O.~Vinyals, and J.~Dean, ``Distilling the knowledge in a neural
  network,'' \emph{CoRR}, vol. abs/1503.02531, 2015.

\bibitem{RomeroBKCGB14}
A.~Romero, N.~Ballas, S.~E. Kahou, A.~Chassang, C.~Gatta, and Y.~Bengio,
  ``Fitnets: Hints for thin deep nets,'' \emph{CoRR}, vol. abs/1412.6550, 2014.

\bibitem{GuptaHM15}
S.~Gupta, J.~Hoffman, and J.~Malik, ``Cross modal distillation for supervision
  transfer,'' in \emph{CVPR}, 2016, pp. 2827--2836.

\bibitem{TzengHDS15}
E.~Tzeng, J.~Hoffman, T.~Darrell, and K.~Saenko, ``Simultaneous deep transfer
  across domains and tasks,'' in \emph{ICCV}, 2015, pp. 4068--4076.

\bibitem{ZhangWWQW16}
B.~Zhang, L.~Wang, Z.~Wang, Y.~Qiao, and H.~Wang, ``Real-time action
  recognition with enhanced motion vector {CNNs},'' in \emph{CVPR}, 2016, pp.
  2718--2726.

\bibitem{BucilaCN06}
C.~Bucila, R.~Caruana, and A.~Niculescu{-}Mizil, ``Model compression,'' in
  \emph{SIGKDD}, 2006, pp. 535--541.

\bibitem{Oliva09}
A.~Oliva, ``Scene perception,'' \emph{Encyclopaedia of Perception}, 2009.

\bibitem{YuZSSX15}
F.~Yu, Y.~Zhang, S.~Song, A.~Seff, and J.~Xiao, ``{LSUN:} construction of a
  large-scale image dataset using deep learning with humans in the loop,''
  \emph{CoRR}, vol. abs/1506.03365, 2015.

\bibitem{WangX16}
L.~Wang, Y.~Xiong, Z.~Wang, Y.~Qiao, D.~Lin, X.~Tang, and L.~{Van Gool},
  ``Temporal segment networks: Towards good practices for deep action
  recognition,'' in \emph{ECCV}, 2016, pp. 20--36.

\bibitem{JiaSDKLGGD14}
Y.~Jia, E.~Shelhamer, J.~Donahue, S.~Karayev, J.~Long, R.~B. Girshick,
  S.~Guadarrama, and T.~Darrell, ``Caffe: Convolutional architecture for fast
  feature embedding,'' \emph{CoRR}, vol. abs/1408.5093.

\bibitem{WangLTL15a}
L.~Wang, C.~Lee, Z.~Tu, and S.~Lazebnik, ``Training deeper convolutional
  networks with deep supervision,'' \emph{CoRR}, vol. abs/1505.02496, 2015.

\bibitem{WangGH015}
L.~Wang, S.~Guo, W.~Huang, and Y.~Qiao, ``Places205-vggnet models for scene
  recognition,'' \emph{CoRR}, vol. abs/1508.01667, 2015.

\bibitem{gao2015deep}
B.~Gao, X.~Wei, J.~Wu, and W.~Lin, ``Deep spatial pyramid: The devil is once
  again in the details,'' \emph{CoRR}, vol. abs/1504.05277, 2015.

\bibitem{XieZYL16}
G.~Xie, X.~Zhang, S.~Yan, and C.~Liu, ``Hybrid {CNN} and dictionary-based
  models for scene recognition and domain adaptation,'' \emph{CoRR}, vol.
  abs/1601.07977, 2016.

\bibitem{Herranz_2016_CVPR}
L.~Herranz, S.~Jiang, and X.~Li, ``Scene recognition with cnns: Objects, scales
  and dataset bias,'' in \emph{CVPR}, 2016, pp. 571--579.

\bibitem{ChangL11}
C.~Chang and C.~Lin, ``{LIBSVM:} {A} library for support vector machines,''
  \emph{{ACM} {TIST}}, vol.~2, no.~3, p.~27, 2011.

\bibitem{YangZC16}
H.~Yang, J.~T. Zhou, and J.~Cai, ``Improving multi-label learning with missing
  labels by structured semantic correlations,'' in \emph{ECCV}, 2016, pp.
  835--851.

\bibitem{YangZZGWC15}
H.~Yang, J.~T. Zhou, Y.~Zhang, B.~Gao, J.~Wu, and J.~Cai, ``Can partial strong
  labels boost multi-label object recognition?'' \emph{CoRR}, vol.
  abs/1504.05843, 2015.

\end{thebibliography}

\end{document}